\documentclass{article}
\usepackage[preprint]{colm2026_conference}
\usepackage{pifont}
\usepackage{multirow}
\usepackage{microtype}
\usepackage{url}
\usepackage{longtable}
\usepackage{booktabs}
\usepackage{lineno}
\usepackage[utf8]{inputenc}
\usepackage{graphicx}
\usepackage{amsmath, amssymb, amsthm}
\usepackage{caption}
\usepackage{subcaption}
\usepackage{float}
\usepackage{enumitem}
\usepackage{algorithm}
\usepackage{algpseudocode}
\usepackage{hyperref}
\usepackage[table]{xcolor}
\usepackage{colortbl}
\usepackage{makecell}
\usepackage{soul}
\usepackage{newunicodechar}
\newunicodechar{−}{-}
\definecolor{darkblue}{rgb}{0,0,0.5}
\definecolor{highlightgreen}{RGB}{144,238,144}
\definecolor{highlightblue}{RGB}{230,240,250}

\hypersetup{colorlinks=true, citecolor=darkblue, linkcolor=darkblue, urlcolor=darkblue}
\usepackage{tcolorbox}
\newtheorem{proposition}{Proposition}

\newtheorem{remark}{Remark}

\algrenewcommand\algorithmicindent{1.0em}

\title{$S^3$: Stratified Scaling Search for Test-Time in Diffusion Language Models}

\author{Ahsan Bilal$^\dagger$ \\
University of Oklahoma \\
\texttt{ahsan.bilal-1@ou.edu}
\And
Muhammad Ahmed Mohsin$^\dagger$ \\
Stanford University \\
\texttt{muahmed@stanford.edu}
\And
Muhammad Umer \\
Stanford University \\
\texttt{mumer@stanford.edu}
\And
Asad Aali \\
Stanford University \\
\texttt{asadaali@stanford.edu}
\And
Muhammad Usman Khanzada \\
University of G\"{o}ttingen \\
\texttt{muhammad.khanzada@stud.uni-goettingen.de}
\And
Muhammad Usman Rafique \\
Zoox \\
\texttt{usman@urafique.com}
\And
Zihao He \\
Meta \\
\texttt{zihaoh@usc.edu}
\And
Emily Fox \\
Stanford University \\
\texttt{ebfox@stanford.edu}
\AND
Dean F. Hougen \\
University of Oklahoma \\
\texttt{hougen@ou.edu}
\AND
{\small $^\dagger$Equal contribution}
}

\begin{document}

\ifcolmsubmission
\linenumbers
\fi

\maketitle

\begin{abstract}
Test-time scaling investigates whether a fixed diffusion language model (DLM) can generate better outputs when given more inference compute, without additional training. However, naive best-of-$K$ sampling is fundamentally limited because it repeatedly draws from the same base diffusion distribution, whose high-probability regions are often misaligned with high-quality outputs. We propose \textbf{$S^3$ (Stratified Scaling Search)}, a classical verifier-guided search method that improves generation by reallocating compute during the denoising process rather than only at the final output stage. At each denoising step, $S^3$ expands multiple candidate trajectories, evaluates them with a lightweight reference-free verifier, and selectively resamples promising candidates while preserving diversity within the search frontier. This procedure effectively approximates a reward-tilted sampling distribution that favors higher-quality outputs while remaining anchored to the model prior. Experiments with LLaDA-8B-Instruct on MATH-500, GSM8K, ARC-Challenge, and TruthfulQA demonstrate that $S^3$ consistently improves performance across benchmarks, achieving the largest gains on mathematical reasoning tasks while leaving the underlying model and decoding schedule unchanged. These results show that classical search over denoising trajectories provides a practical mechanism for test-time scaling in DLMs.
\end{abstract}

\section{Introduction} \label{sec:intro}

Test-time scaling addresses a fundamental question: given a fixed model and additional compute at inference, how much can performance improve? For autoregressive language models, the answer depends strongly on how compute is spent: chain-of-thought reasoning, best-of-$K$ sampling, and tree search have all been shown to significantly enhance performance \citep{brown2024largelanguagemonkeysscaling,snell2024scaling}. Diffusion language models (DLMs) offer a structurally distinct opportunity. Recent diffusion LLMs already demonstrate competitive, and sometimes stronger, performance than similarly sized autoregressive models \citep{zhao2025d1}.

For autoregressive models, inference-time compute has been allocated across token-level search strategies such as beam search, MCTS, and best-of-$K$ sampling \citep{brown2024largelanguagemonkeysscaling, snell2024scaling}, each exploiting the sequential left-to-right structure of generation. This opportunity extends naturally to DLMs: because generation proceeds through iterative denoising over $T$ steps, decoding forms a stochastic sequential process in which partial states are gradually refined into a final sequence, enabling multiple possible denoising trajectories at each step \citep{kim2025train}. Unlike autoregressive decoding, however, standard diffusion decoding samples only a single trajectory \citep{kim2025train, nie2024scaling}, leaving this structure unexploited. A common inference-time strategy adapted from autoregressive models is best-of-$K$ sampling (BoK), which is fundamentally limited by the fact that increasing the number of samples does not change the underlying distribution from which they are drawn, a limitation that applies equally to both autoregressive and diffusion settings, as we formalize in Section~\ref{sec:mismatch}.

To address this limitation, we analyze the inference-time objective from a distributional perspective and show in Section~\ref{sec:gibbs} that the optimal distribution under a KL constraint relative to the model prior is a reward-tilted Gibbs distribution. Motivated by this objective, we introduce \textbf{$S^3$ (Stratified Scaling Search)}, an inference-time procedure that maintains a population of partial denoising trajectories and allocates computation toward trajectories identified as promising according to verifier-based look-ahead estimates, as described in Section~\ref{sec:method}.

\paragraph{Contributions}
(1) We identify a \textbf{density-quality mismatch} in DLMs, where high-probability regions of $p_0(x)$ are misaligned with verifier rewards, limiting naive best-of-$K$ sampling. (2) We show that the optimal inference-time target under a KL constraint is a \textbf{reward-tilted Gibbs distribution} $\tilde{p}_0(x) \propto p_0(x)\exp(\tau f(x))$, which shifts probability mass toward high-scoring outputs while remaining anchored to the model prior. (3) We propose \textbf{$S^3$}, a verifier-guided particle search over denoising trajectories that, without retraining, employs a lightweight verifier requiring no ground-truth labels and no LLM-as-a-judge, and improves accuracy from 25.60\% to 30.20\% on MATH-500, 68.16\% to 70.21\% on GSM8K, and 46.49\% to 49.57\% on TruthfulQA, while achieving competitive performance on ARC-Challenge (76.11\% to 77.86\%), where best-of-$K$ can outperform under coarse block lengths (see Section~\ref{sec:eval}).

\begin{figure*}[t]
\centering

\begin{subfigure}[t]{0.32\textwidth}
\centering
\includegraphics[width=\linewidth]{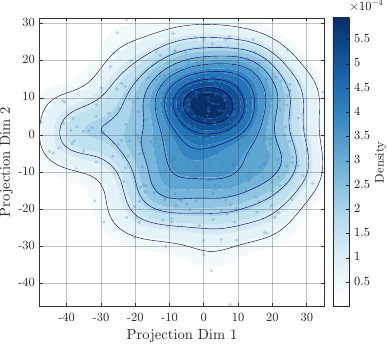}
\caption{Base distribution $p_0$.}
\label{fig:1a}
\end{subfigure}\hfill
\begin{subfigure}[t]{0.32\textwidth}
\centering
\includegraphics[width=\linewidth]{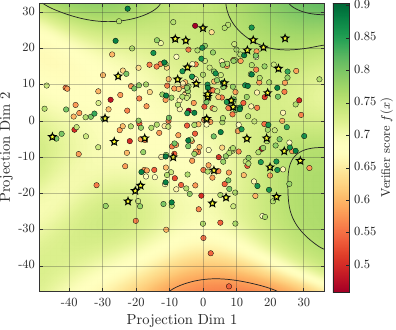}
\caption{Verifier landscape $f(x)$.}
\label{fig:1b}
\end{subfigure}\hfill
\begin{subfigure}[t]{0.32\textwidth}
\centering
\includegraphics[width=\linewidth]{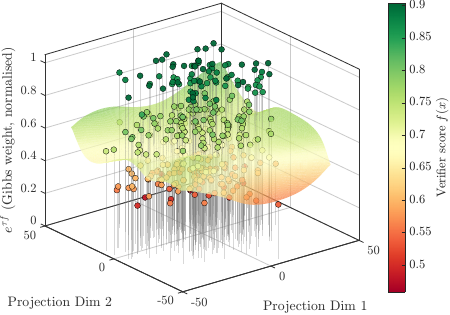}
\caption{Reward-tilted target $\tilde p_0$.}
\label{fig:1c}
\end{subfigure}

\begin{subfigure}[t]{0.32\textwidth}
\centering
\includegraphics[width=\linewidth]{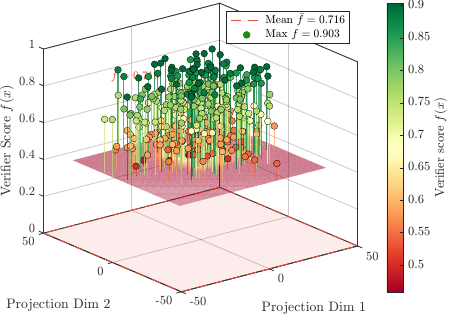}
\caption{Direct sampling from $p_0$.}
\label{fig:1d}
\end{subfigure}\hfill
\begin{subfigure}[t]{0.32\textwidth}
\centering
\includegraphics[width=\linewidth]{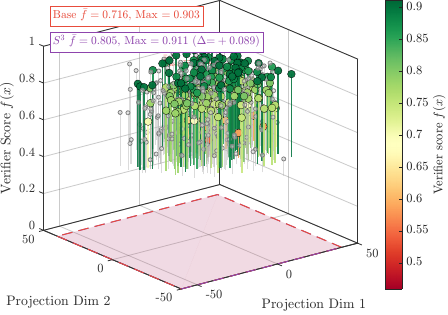}
\caption{$S^3$ trajectory search.}
\label{fig:1e}
\end{subfigure}\hfill
\begin{subfigure}[t]{0.32\textwidth}
\centering
\includegraphics[width=\linewidth]{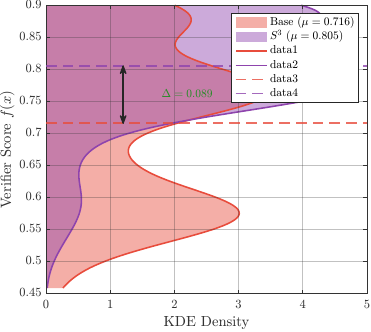}
\caption{Score distribution shift.}
\label{fig:1f}
\end{subfigure}

\caption{\textbf{Density--quality mismatch and trajectory search under $S^3$ for MATH-500}. Panels (a)-(c) illustrate the distributional mismatch between $p_0$, the verifier landscape $f(x)$, and the reward-tilted target $\tilde{p}_0$. Panels (d)-(f) show the corresponding inference behavior of direct sampling versus $S^3$ trajectory search and the resulting shift in verifier score distribution.}
\label{fig:distribution_geometry}
\end{figure*}

\section{Preliminaries} \label{sec:motivation}

DLMs generate text through a reverse denoising process $x_T \rightarrow x_{T-1} \rightarrow \cdots \rightarrow x_0$, where $x_T$ is an initial fully masked sequence sampled from a simple prior distribution, and $x_0$ is the final decoded sequence. Under a fixed decoding schedule, this process induces a base output distribution $p_0$ over complete sequences. 

\subsection{The Density-Quality Mismatch} \label{sec:mismatch}

Let $p_0(x)$ denote the base output distribution and let $f(x)$ denote the verifier score. The expected quality under direct sampling is $\mathcal{Q}(p_0) = \mathbb{E}_{x\sim p_0}[f(x)]$. As shown in Figure~\ref{fig:1a}, the base model distribution $p_0$ often places probability mass on regions that are not aligned with verifier rewards. Meanwhile, the verifier landscape in Figure~\ref{fig:1b} highlights that high-quality outputs may lie in sparse regions of the space. The reward-tilted target distribution illustrated in Figure~\ref{fig:1c}, therefore, concentrates probability mass in areas with higher verifier scores. While such misalignment has been noted in autoregressive settings \citep{snell2024scaling, brown2024largelanguagemonkeysscaling}, we formalize it for DLMs as the density-quality mismatch and show in Section~\ref{sec:gibbs} that the Gibbs tilt \citep{donsker1975asymptotic} provides the principled correction. A natural remedy is best-of-$K$ sampling, which, however, remains fundamentally limited as we show next.
\begin{remark}
\label{rem:best_of_k}
Suppose we draw $K$ i.i.d.\ samples $x_1,\dots,x_K \sim p_0$ and return the highest-scoring one. Under mild regularity assumptions on $f$ and $p_0$, $\mathbb{E}\!\left[\max_{k \le K} f(x_k)\right] \le \mathcal{Q}(p_0) + \sigma\sqrt{2\ln K}$, where $\sigma^2 = \mathrm{Var}_{p_0}[f]$. This bound follows from standard concentration inequalities for maxima of sub-Gaussian random variables \citep{concentration_inequality}. The improvement, therefore, grows only logarithmically with $K$, underscoring the need for more efficient inference-time compute allocation strategies.
\end{remark}

\subsection{The Optimal Inference-Time Target} \label{sec:gibbs}

A principled objective is to maximize the expected verifier reward while remaining close to the base model distribution: $\max_{\tilde p} \; \mathbb{E}_{\tilde p}[f(x)]$ subject to $D_{\mathrm{KL}}(\tilde p \,\|\, p_0) \le \delta$.

\begin{proposition}[Optimal reward-tilted distribution]
\label{prop:gibbs_optimal}
The unique maximizer of the above objective is the Gibbs distribution
\begin{equation}
\tilde{p}_0(x_0;\tau) \;\propto\; p_0(x_0)\,e^{\tau f(x_0)}
\label{eq:gibbs}
\end{equation}
\end{proposition}
This result follows from standard variational formulations of KL-constrained optimization and exponential tilting of probability measures \citep{donsker1975asymptotic}. As illustrated in Figures~\ref{fig:1d}--\ref{fig:1f}, direct sampling from $p_0$ remains concentrated around the model prior. Remark~\ref{rem:best_of_k} shows that filtering outputs from $p_0$ yields only logarithmic gains in $K$, making output-level selection fundamentally insufficient. Approximating $\tilde{p}_0$ instead requires \emph{reshaping} the sampling distribution by exploiting the sequential denoising structure, formalized as a tractable look-ahead approximation in Level~3 of Section~\ref{sec:method}.

\begin{figure}[t]
\centering
\includegraphics[width=\linewidth]{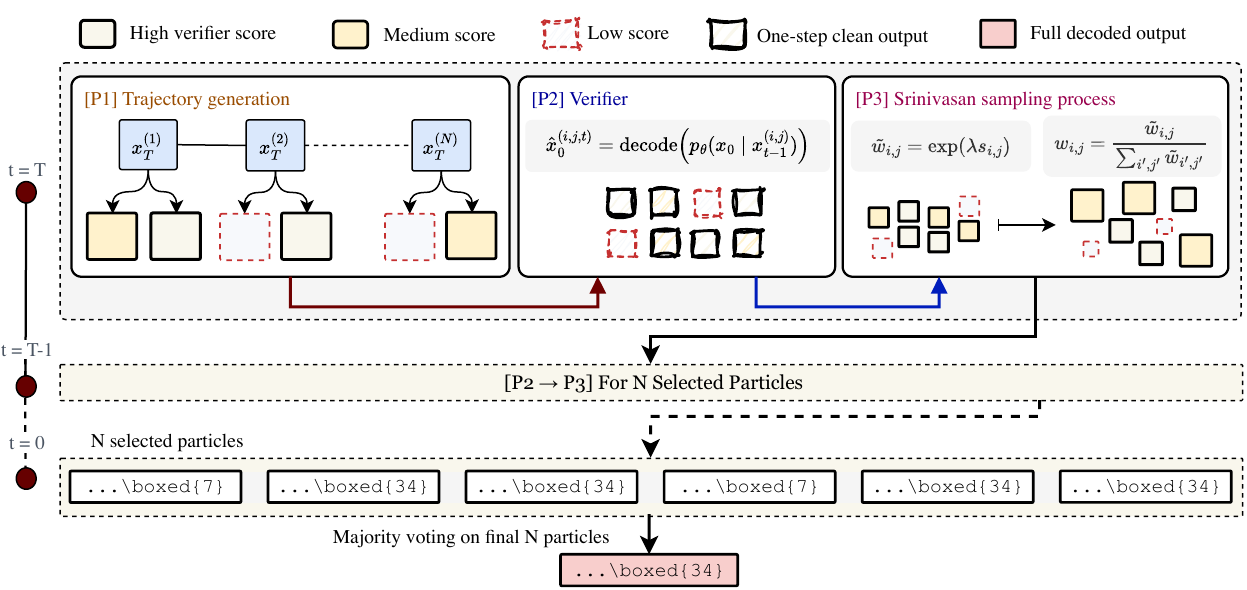}
\caption{\textbf{$S^3$:} $N$ particles are initialized at $t{=}T$ [P1], then at each step expanded into $Nb$ candidates, scored via one-step clean predictions $\hat{x}_0^{(i,j,t)}$ [P2], and resampled to $N$ particles via SSP [P3]. Final output is selected by majority voting at $t{=}0$.}
\label{fig:bfs_ssp_pipeline}
\end{figure}

\section{Methodology}
\label{sec:method}

Section~\ref{sec:motivation} established that direct sampling from the base diffusion distribution $p_0$ is limited by a density--quality mismatch, while the ideal inference-time target is the Gibbs-tilted distribution in Eq.~\eqref{eq:gibbs}. We approximate this target in practice using \textbf{$S^3$}, an inference-time search procedure over denoising trajectories that requires no retraining and operates on a fixed diffusion decoding schedule, developed across three levels: (1) the \emph{exact} terminal target, (2) its exact sequential realization, and (3) the tractable approximation that $S^3$ implements as in Figure~\ref{fig:bfs_ssp_pipeline}.

\subsection{Diffusion decoding as a path-space problem}
\label{subsec:path_space}

Let $x$ denote the input prompt and $L$ the output sequence length. Under a fixed decoding schedule $e=(\mathcal{K},\pi,T)$, the DLM defines a reverse denoising chain from the fully masked initial state $x_T \in \mathcal{X}^L$ to the final decoded sequence $x_0 \in \mathcal{X}^L$, where $\mathcal{K}$ is the block length controlling how many positions are unmasked per step, $\pi$ is the masking update policy, and $T = \lceil L/\mathcal{K} \rceil$ is the total number of denoising steps. This formulation is agnostic to the specific model instantiation, discrete \citep{do2025discrete}, continuous, or energy-based \citep{xu2024energy} as long as a stochastic reverse transition $p_\theta(x_{t-1}\mid x_t)$ is defined \citep{nie2025large}.

At each step $t$, the schedule specifies a set of $\mathcal{K}$ updatable positions $U_t \subseteq \{1,\dots,L\}$, where the model samples $x_{t-1} \sim p_\theta(\cdot \mid x_t,\,e)$ while holding all positions outside $U_t$ fixed, i.e., $x_{t-1}[i] = x_t[i]$ for all $i \notin U_t$, and $x_t[i]$ denotes the token at position $i$ of the partial state $x_t$. The collection of these reverse transitions, chained over all $T$ steps, defines a \emph{base path measure} over complete denoising trajectories $x_{0:T} = (x_0, x_1, \dots, x_T)$ \citep{nie2025large},
\begin{equation}
p(x_{0:T})
  \;=\;
  p_T(x_T)
  \prod_{t=1}^{T}
  p_\theta(x_{t-1}\mid x_t,\,e),
  \label{eq:base_path}
\end{equation}
where $p_T$ is the prior over the fully masked initial state $x_T$, each factor $p_\theta(x_{t-1}\mid x_t,e)$ is the reverse transition kernel at step $t$, and the terminal marginal $p_0(x_0) = \sum_{x_{1:T}} p(x_{0:T})$ recovers the base output distribution from Section~\ref{sec:motivation}; here, the sum is over all intermediate trajectories $x_{1:T}$ consistent with the fixed endpoints $x_0$ and $x_T$. 

\paragraph{Level 1: The exact terminal target.}
As established in Section~\ref{sec:gibbs}, the optimal inference-time distribution under a KL constraint is the Gibbs tilt distribution defined in Eq.~\eqref{eq:gibbs}, where $f:\mathcal{X}^L\to[0,1]$ is the verifier and $\tau > 0$ is the reward temperature. Extending this reward tilt from final outputs to the full denoising trajectory gives the \emph{target path measure}
\begin{equation}
\tilde{p}\!\left(x_{0:T}\right)
  \;\propto\;
  p\!\left(x_{0:T}\right)\,e^{\tau f(x_0)},
  \label{eq:tilted_path}
\end{equation}
whose terminal marginal is exactly the desired reward-tilted distribution $\tilde{p}_0$ in Eq.~\eqref{eq:gibbs}.

\paragraph{Level 2: The exact sequential realization.}
Sampling from $\tilde{p}$ exactly requires \emph{twisting} the reverse transition kernels, i.e., the learned conditional distributions $p_\theta(x_{t-1} \mid x_t, e)$ that define the probability of transitioning from state $x_t$ to $x_{t-1}$ at every denoising step. Define the \emph{backward information function} at step $t$ as
\begin{equation}
h_t(x_t)
  \;:=\;
  \mathbb{E}_{p}\!\left[e^{\tau f(x_0)} \mid x_t = x_t\right],
  \label{eq:backward_info}
\end{equation}
where $x_0$ is the terminal output distributed according to $p_0$. The function $h_t$ encodes all expected future reward reachable from the current partial state $x_t$, satisfying boundary conditions $h_T(x_T) = \tilde{p}_0 / p_0$ up to a normalizing constant, and $h_0(x_0) = e^{\tau f(x_0)}$. The marginal of $\tilde{p}$ at step $t$ factorizes as $\tilde{p}_t(x_t) \;\propto\; p_t(x_t)\,h_t(x_t)$, where $p_t$ is the $t$-step marginal of $p$, and the exact \emph{twisted reverse kernel} that realizes $\tilde{p}$ sequentially is
\begin{equation}
  \tilde{p}_\theta\!\left(x_{t-1}\mid x_t\right)
  \;=\;
  p_\theta\!\left(x_{t-1}\mid x_t,\,e\right)
  \cdot
  \frac{h_{t-1}(x_{t-1})}{h_t(x_t)}.
  \label{eq:twisted_kernel}
\end{equation}
This is the Doob $h$-transform of the base kernel \citep{whiteley2014twisted}, where the base transition is weighted by the ratio of future reward expectations at the child state relative to the parent state. Formally, $G_t(x_t,x_{t-1}) := h_{t-1}(x_{t-1})/h_t(x_t)$ is the \emph{incremental potential} at step $t$, and a Sequential Monte Carlo (SMC) sampler using these potentials as importance weights yields an exact, unbiased particle approximation of $\tilde{p}$ \citep{smith2013sequential,neal2001annealed}. The auxiliary particle filter of \citet{pitt1999filtering} can be understood as approximating this look-ahead weighting scheme precisely via a predictive approximation of the backward information function.

\paragraph{Level 3: The $S^3$ approximation.}
The backward information functions $h_t$ defined in Eq.~\eqref{eq:backward_info} are intractable: computing them exactly requires marginalizing over all denoising trajectories from step $t$ to $0$ under the base path measure, which is as expensive as running the full reverse process from every candidate state. $S^3$ replaces these intractable quantities with a tractable surrogate derived from the diffusion model's own one-step clean prediction, described next.

\subsection{Trajectory search with look-ahead scoring}
\label{subsec:trajectory_search}

The reverse process in Eq.~\eqref{eq:base_path} naturally induces a search tree over partial states. We maintain a population of $N$ partial trajectories $\{x_t^{(i)}\}_{i=1}^N$ and expand each particle into $b$ candidate successors by sampling $x_{t-1}^{(i,j)} \sim p_\theta\!\left(\cdot \mid x_t^{(i)},\,e\right)$ for $j=1,\dots,b$, producing an expanded frontier of $Nb$ candidate partial trajectories $\{x_{t-1}^{(i,j)}\}$. Since $x_0$ is unavailable at intermediate steps, evaluating $h_{t-1}(x_{t-1}^{(i,j)})$ directly is intractable. We instead approximate it via the model's one-step clean prediction,
\begin{equation}
  \hat{x}_0^{(i,j,t)}
  \;=\;
  \mathrm{decode}\!\Bigl(p_\theta\!\left(x_0 \mid x_{t-1}^{(i,j)}\right)\Bigr),
  \label{eq:clean_pred}
\end{equation}
where $\mathrm{decode}(\cdot)$ denotes argmax decoding, and apply the verifier to obtain a look-ahead score
\begin{equation}
  s_{i,j,t} \;=\; f\!\left(\hat{x}_0^{(i,j,t)}\right) \;\in\; [0,1],
  \label{eq:lookahead_score}
\end{equation}
which estimates how rewarding the trajectory through $x_{t-1}^{(i,j)}$ is likely to be at termination. This defines the approximate backward information function
\begin{equation}
  \hat{h}_{t-1}\!\left(x_{t-1}^{(i,j)}\right)
  \;:=\;
  \exp\!\left\{\lambda\, s_{i,j,t}\right\}
  \;\approx\;
  h_{t-1}\!\left(x_{t-1}^{(i,j)}\right),
  \label{eq:approx_twist}
\end{equation}
where $\lambda > 0$ is an inverse-temperature hyperparameter, making $s_{i,j,t}$ a tractable surrogate for $\tfrac{1}{\tau}\log h_{t-1}(x_{t-1}^{(i,j)})$. This instantiates for DLM decoding the same look-ahead principle underlying the auxiliary particle filter \citep{pitt1999filtering} and twisted particle filters \citep{whiteley2014twisted}: the model's one-step denoising prediction plays the role of the look-ahead, and the verifier provides the reward signal.

\begin{algorithm}[t]
\caption{Single-Schedule Verifier-Guided Particle Search ($S^3$)}
\label{alg:s3}
\begin{algorithmic}[1]
\Require Diffusion model $p_\theta$, verifier $f$, prompt $x$, fixed schedule $e=(\mathcal{K},\pi,T)$, particle count $N$, branching factor $b$, temperature $\lambda>0$
\Ensure Best decoded output $y^\star$
\State \textbf{Initialize:} $x_T^{(i)} \leftarrow \textsc{FullMask}(x)$ for $i=1,\dots,N$
\Comment{all output positions masked}
\For{$t = T, T-1, \dots, 1$}
    \State Determine updatable positions $U_t \subseteq \{1,\dots,L\}$ from schedule policy $\pi$ \Comment{$|\,U_t\,| = \mathcal{K}$}
    \For{each particle $i \in \{1,\dots,N\}$}
        \For{each child $j \in \{1,\dots,b\}$}
            \State $x_{t-1}^{(i,j)} \sim p_\theta(\cdot \mid x_t^{(i)}, e)$ \Comment{update only $\mathcal{K}$ positions in $U_t$}
            \State $\hat{x}_0^{(i,j,t)} \leftarrow \mathrm{decode}\!\left(p_\theta(x_0 \mid x_{t-1}^{(i,j)})\right)$
            \Comment{clean prediction, Eq.~\eqref{eq:clean_pred}}
            \State $s_{i,j,t} \leftarrow f\!\left(\hat{x}_0^{(i,j,t)}\right)$
            \Comment{look-ahead verifier score, Eq.~\eqref{eq:lookahead_score}}
        \EndFor
    \EndFor
    \State $\tilde{w}_{i,j,t} \leftarrow \exp(\lambda s_{i,j,t})$ for all $i,j$
    \Comment{approximate twist weight, Eq.~\eqref{eq:approx_twist}}
    \State $w_{i,j,t} \leftarrow \tilde{w}_{i,j,t}\big/\sum_{i',j'} \tilde{w}_{i',j',t}$
    \Comment{normalize over all $N\cdot b$ children}
    \State $\xi_{i,j,t} \leftarrow N\, w_{i,j,t}$ for all $i,j$
    \Comment{expected offspring counts}
    \State $\{x_{t-1}^{(i)}\}_{i=1}^{N} \leftarrow \textsc{SSPResample}\!\left(\{x_{t-1}^{(i,j)}\}, \{\xi_{i,j,t}\}, N\right)$
    \Comment{low-variance dependent-rounding resampling}
\EndFor
\State \Return $y^\star \leftarrow \mathrm{MajorityVote}\!\left(\{x_0^{(i)}\}_{i=1}^N\right)$ \Comment{break ties by lowest NLL under the base model}
\end{algorithmic}
\end{algorithm}

\subsection{Verifier-weighted resampling with Srinivasan Sampling Process (SSP)}
\label{subsec:ssp}

To bias the search toward higher-quality trajectories while preserving particle diversity, we convert look-ahead scores from Eq.~\eqref{eq:lookahead_score} into normalized importance weights. For each expanded candidate $x_{t-1}^{(i,j)}$ in the frontier of $Nb$ children, the unnormalized weight $\tilde{w}_{i,j,t} = \exp\!\left(\lambda\, s_{i,j,t}\right)$ and normalized weight $w_{i,j,t} = \tilde{w}_{i,j,t} / \sum_{i',j'} \tilde{w}_{i',j',t}$ over all $Nb$ candidates, where larger $\lambda$ concentrates mass on high-scoring trajectories and smaller $\lambda$ retains a more diffuse particle distribution.

\begin{remark}
\label{rem:approx_weights}
The per-step weights $\tilde{w}_{i,j,t}$ are inspired by the incremental potential $G_t(x_t,x_{t-1}) = h_{t-1}(x_{t-1})/h_t(x_t)$ from Eq.~\eqref{eq:twisted_kernel}, but do not implement the exact sequential importance sampling correction for $\tilde{p}$, as doing so would require computing the ratio of consecutive backward information functions at consecutive states, which is intractable. $S^3$ instead uses the surrogate $\hat{h}_{t-1}$ from Eq.~\eqref{eq:approx_twist} as a stand-alone potential without dividing by $\hat{h}_t(x_t^{(i)})$, and should therefore be interpreted as a verifier-guided approximate twisting scheme rather than an unbiased particle filter for $\tilde{p}$.
\end{remark}

The expected offspring allocation $\xi_{i,j,t} = N\, w_{i,j,t}$ with $\sum_{i,j}\xi_{i,j,t} = N$ is converted into exact integer counts using the \textbf{Srinivasan Sampling Process (SSP)} \citep{srinivasan2001distributions}, a low-variance dependent-rounding procedure with $n_{i,j,t}\in\mathbb{Z}_{\ge 0}$ and $\sum_{i,j} n_{i,j,t}=N$. SSP retains stochasticity in the resampling step, which is critical because deterministic top-$k$ pruning leads to mode collapse within the search tree, whereas stochastic resampling preserves multiple plausible denoising trajectories.

Each denoising step thus applies an \emph{expand--score--resample} update: \textbf{expansion} proposes $Nb$ continuations under $p_\theta(\cdot\mid x_t,e)$; \textbf{scoring} evaluates clean-prediction quality $\hat{x}_0$ through the verifier; and \textbf{SSP resampling} reallocates the particle budget toward more promising trajectory regions. Repeating over $T$ steps progressively shifts the population toward the reward-tilted distribution in Eq.~\eqref{eq:tilted_path}; see Appendix~\ref{app:dist_shift} for an empirical illustration.

\subsection{Verifier, compute budget, and final output selection}
\label{subsec:verifier_budget_output}

We use a lightweight ground-truth-free composite verifier scoring candidate outputs via intrinsic signals from the generated text, combining structural validity, arithmetic consistency, answer reachability, model confidence, and degeneracy penalties into a single scalar, $f(x) \;=\; \sum_{k=1}^{5} \alpha_k\, s_k(x) \;+\; \alpha_c\, s_{\text{constraint}}(x)$ where $s_k(x)$ for $k=1,\dots,5$ measure structural completeness, arithmetic consistency, answer reachability, model confidence, and non-degeneracy respectively; $s_{\text{constraint}}(x)$ captures task-specific constraint satisfaction; and $\sum_{k=1}^{5}\alpha_k + \alpha_c = 1$. Full details and dataset-specific weight profiles are given in Appendix~\ref{app:verifier}.

The computational cost is explicit and tunable: each denoising step evaluates $Nb$ transition proposals and $Nb$ clean-prediction verifier scores, yielding approximately $TNb$ total evaluations over $T$ steps, so $N$, $b$, and $\lambda$ directly control the compute--quality trade-off. After the final step, let $\{x_0^{(i)}\}_{i=1}^N$ denote the surviving decoded particles. We extract the final answer from each particle, select the majority answer, and break ties using the lowest negative log-likelihood (NLL) under the base model. The returned output $y^\star$ is the representative decoded sequence corresponding to this selected answer; full latency and computational cost analysis are provided in Appendix~\ref{app:ablation_cost}.
\begin{table}
\begin{center}
\caption{Accuracy (\%) of standard diffusion decoding, Best-of-$K$ ($K{=}8$), and $S^3$ ($N{=}4$, $b{=}2$, $\mathcal{K}{=}64$) on LLaDA-8B-Instruct.}
\label{tab:results}
\begin{tabular}{lccc}
\toprule
\rowcolor{gray!15}
\multicolumn{1}{c}{\bf Dataset} &
\multicolumn{1}{c}{\bf Baseline Diffusion} &
\multicolumn{1}{c}{\bf Best-of-$K$} &
\multicolumn{1}{c}{\bf $S^3$} \\
\midrule
GSM-8K        & 68.16 & 69.56 & 70.21 \\
MATH-500      & 25.60 & 28.20 & 30.20 \\
TruthfulQA    & 46.49 & 49.36 & 49.57 \\
ARC-Challenge & 76.11 & 79.30 & 77.86 \\
\bottomrule
\end{tabular}
\end{center}
\end{table}

\begin{figure*}[t]
\centering
\begin{subfigure}[b]{0.23\linewidth}
\centering
\includegraphics[width=\linewidth]{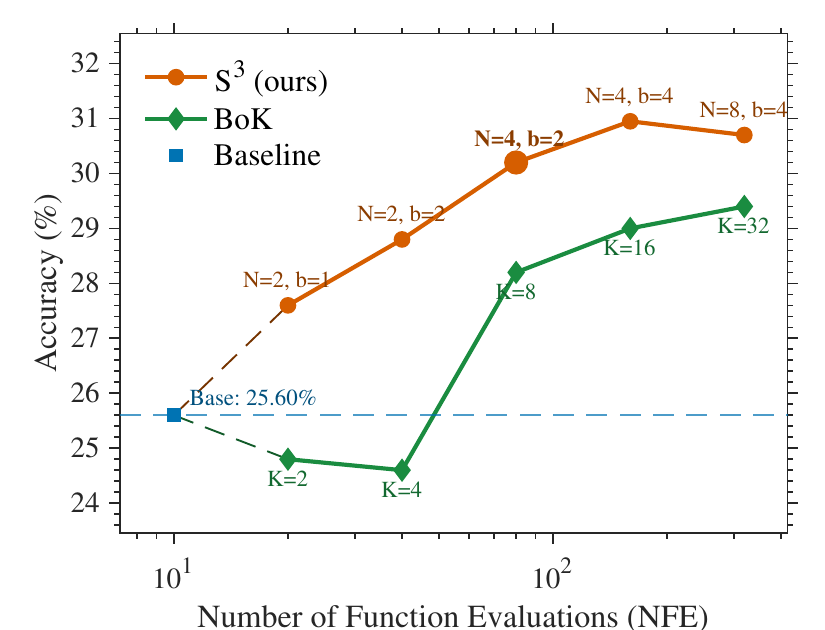}
\caption{MATH-500}
\end{subfigure}
\hfill
\begin{subfigure}[b]{0.23\linewidth}
\centering
\includegraphics[width=\linewidth]{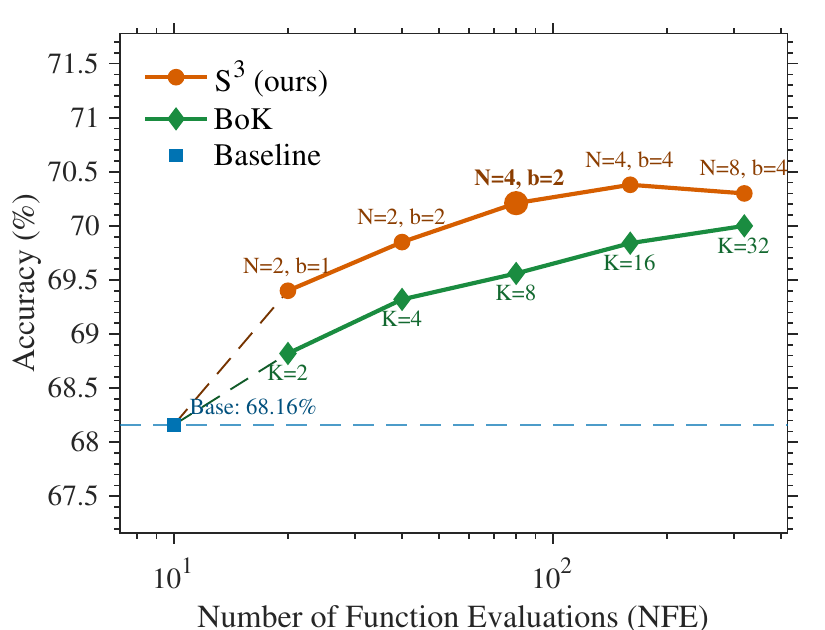}
\caption{GSM8K}
\end{subfigure}
\hfill
\begin{subfigure}[b]{0.23\linewidth}
\centering
\includegraphics[width=\linewidth]{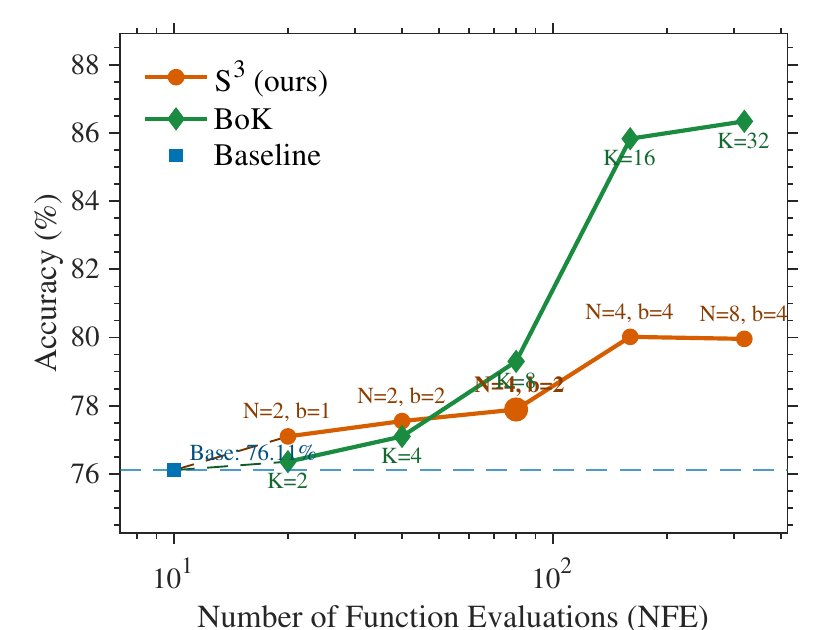}
\caption{ARC-C}
\end{subfigure}
\hfill
\begin{subfigure}[b]{0.23\linewidth}
\centering
\includegraphics[width=\linewidth]{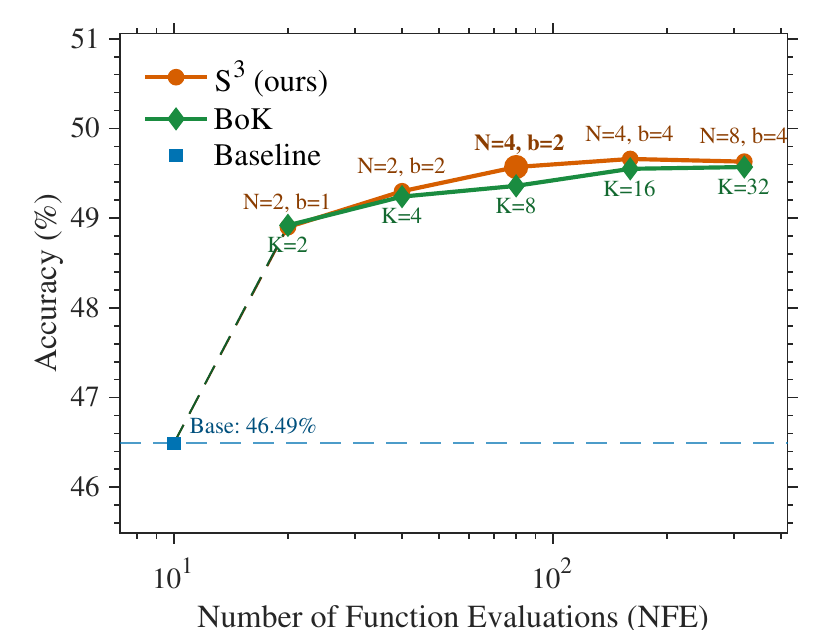}
\caption{TruthfulQA}
\end{subfigure}
\begin{subfigure}[b]{0.23\linewidth}
\centering
\includegraphics[width=\linewidth]{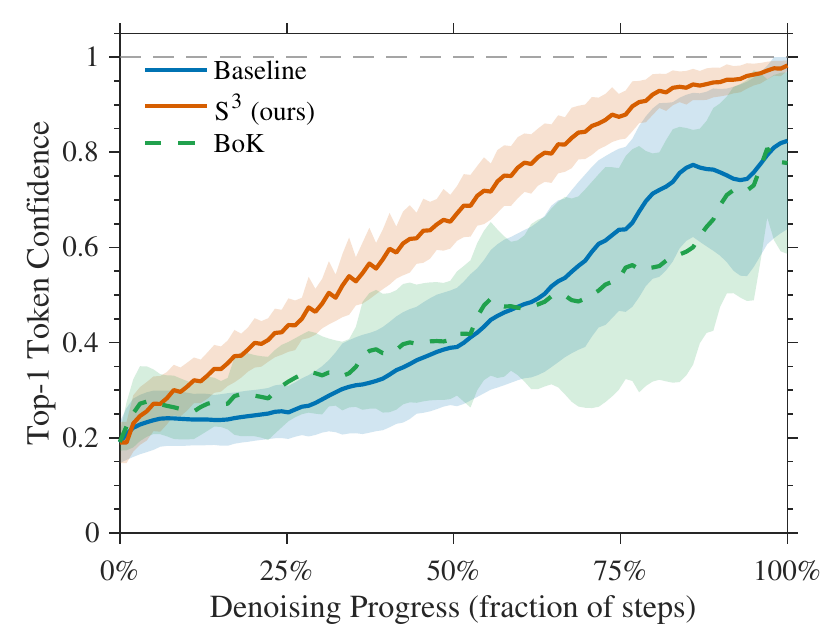}
\caption{MATH-500}
\end{subfigure}
\hfill
\begin{subfigure}[b]{0.23\linewidth}
\centering
\includegraphics[width=\linewidth]{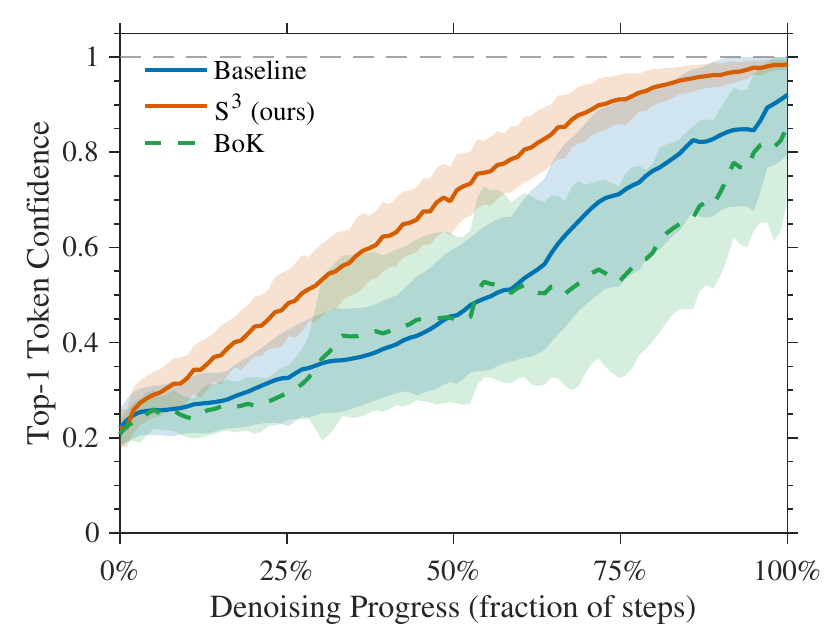}
\caption{GSM8K}
\end{subfigure}
\hfill
\begin{subfigure}[b]{0.23\linewidth}
\centering
\includegraphics[width=\linewidth]{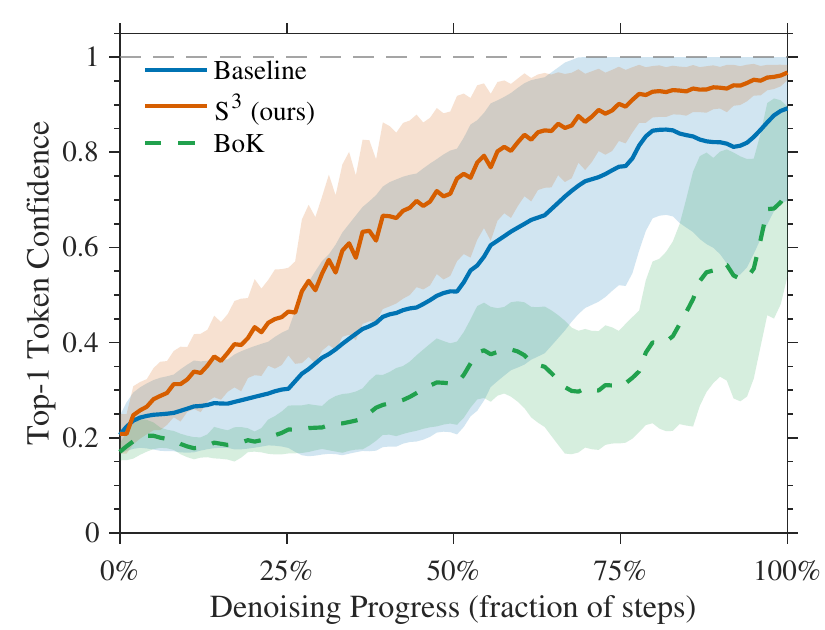}
\caption{ARC-C}
\end{subfigure}
\hfill
\begin{subfigure}[b]{0.23\linewidth}
\centering
\includegraphics[width=\linewidth]{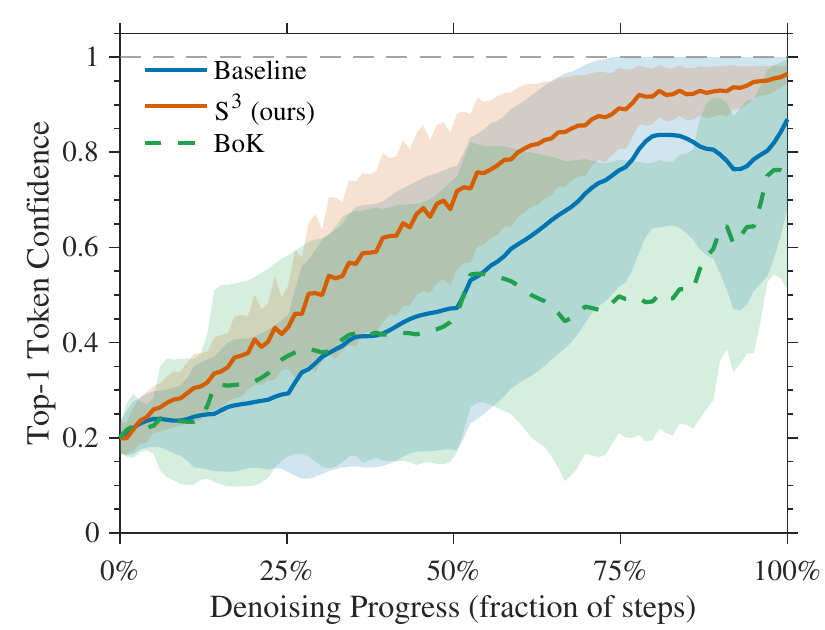}
\caption{TruthfulQA}
\end{subfigure}
\caption{\textbf{Inference-time scaling with $S^3$ across datasets.} Top row: accuracy vs.\ compute ($\mathrm{NFE} = \text{steps} \times N \times b$) across multiple $(N,b)$ settings. Bottom row: mean top-1 token confidence over denoising progress. Curves shown for Baseline, $S^3$ ($N{=}4$, $b{=}2$, $\lambda{=}1.0$), and BoK ($K{=}8$).}
\label{fig:multi_dataset_scaling}
\end{figure*}

\begin{figure*}[t]
  \centering
  %
  \begin{subfigure}[b]{0.245\linewidth}
    \centering
    \includegraphics[width=\linewidth]{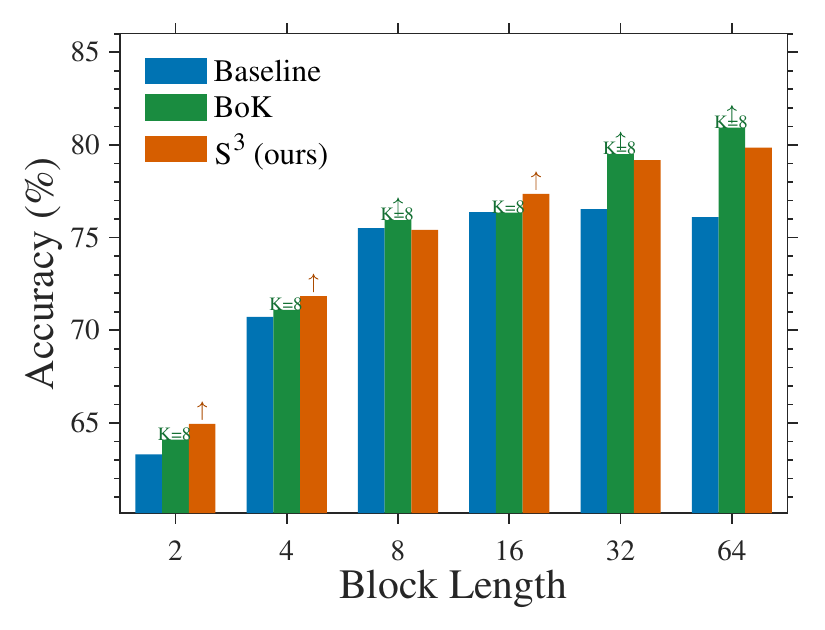}
    \caption{ARC-Challenge}
    \label{fig:block_arc}
  \end{subfigure}%
  \hfill
  \begin{subfigure}[b]{0.245\linewidth}
    \centering
    \includegraphics[width=\linewidth]{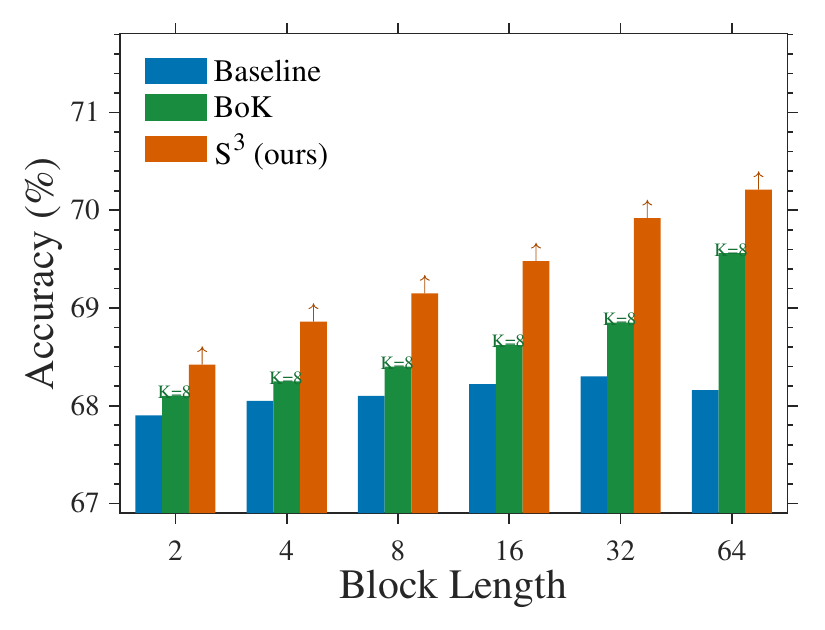}
    \caption{GSM8K}
    \label{fig:block_gsm8k}
  \end{subfigure}%
  \hfill
  \begin{subfigure}[b]{0.245\linewidth}
    \centering
    \includegraphics[width=\linewidth]{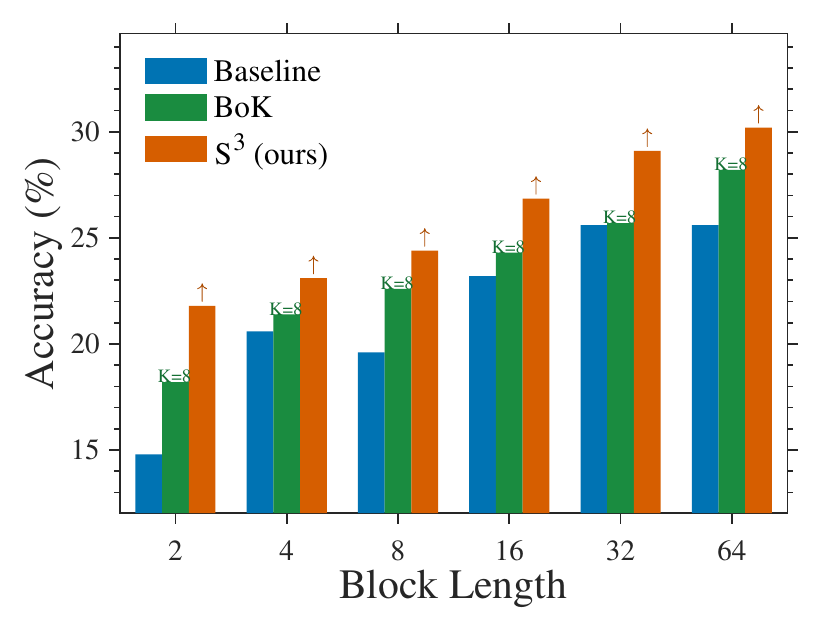}
    \caption{MATH-500}
    \label{fig:block_math}
  \end{subfigure}%
  \hfill
  \begin{subfigure}[b]{0.245\linewidth}
    \centering
    \includegraphics[width=\linewidth]{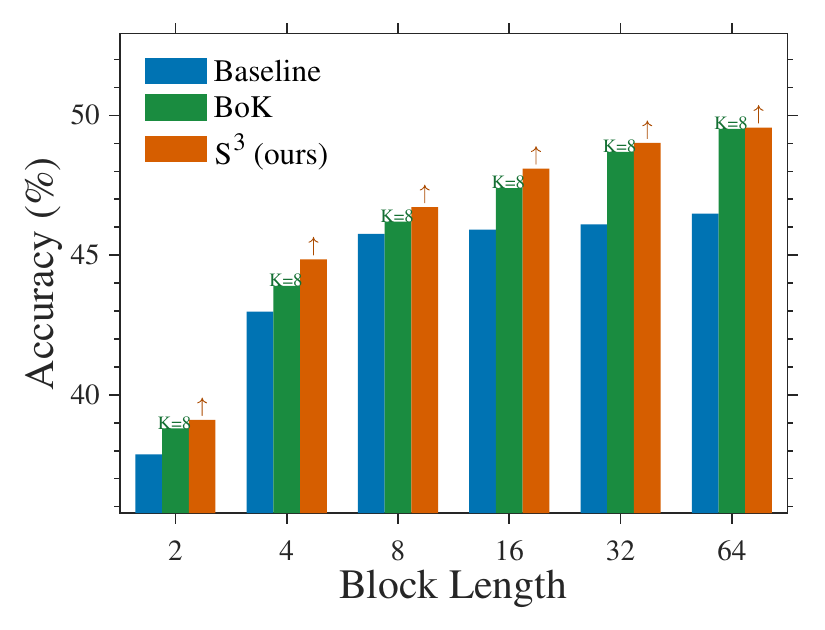}
    \caption{TruthfulQA}
    \label{fig:block_tqa}
  \end{subfigure}
\caption{\textbf{Accuracy (\%) across block lengths $\mathcal{K} \in \{2,4,8,16,32,64\}$ on four benchmarks.} Results use LLaDA-8B-Instruct with $N{=}4$, $b{=}2$, and $K{=}8$. Orange bars denote $S^3$, blue bars denote Baseline, and green bars denote BoK. An upward arrow ($\uparrow$) indicates configurations where the leading method improves over the Baseline.}

  \label{fig:block_length_comparison}
\end{figure*}

\section{Evaluation and Results}
\label{sec:eval}

\paragraph{Experimental Setup}
We evaluate $S^3$ on four benchmarks encompassing mathematical reasoning, scientific reasoning, and factual accuracy: \textbf{GSM8K} \citep{cobbe2021training}, \textbf{MATH-500} \citep{hendrycks2021measuring}, \textbf{TruthfulQA} \citep{lin2022truthfulqa}, and \textbf{ARC-Challenge} \citep{clark2018think}, leveraging \texttt{LLaDA-8B-Instruct} under its default denoising schedule. We follow the evaluation protocol of \citet{lee2025test} for preprocessing and answer normalization. The baseline is standard single-trajectory diffusion decoding from $p_0$; full experimental details are provided in Appendix~\ref{app:exp_details}.

\paragraph{Benchmark Performance}
Table~\ref{tab:results} reports accuracy for the baseline, Best-of-$K$ ($K{=}8$), and $S^3$ ($N{=}4$, $b{=}2$). $S^3$ consistently outperforms both baselines across all benchmarks. We use $(N{=}4, b{=}2)$ as a representative compute point ($\mathrm{NFE} \propto T \cdot N \cdot b$) without dataset-specific tuning; results remain consistent across all $(N,b)$ configurations (Figure~\ref{fig:multi_dataset_scaling}). The most pronounced improvement is observed on MATH-500 (+4.60 pp over baseline, +2.00 pp over BoK), suggesting that verifier-guided trajectory search is particularly effective on tasks requiring multi-step reasoning where intermediate denoising decisions compound. Gains on TruthfulQA (+3.08 pp over baseline) further demonstrate that the density-quality mismatch is not specific to mathematical reasoning. On ARC-Challenge, $S^3$ improves over the baseline (+1.75 pp) but underperforms BoK at $\mathcal{K}{=}64$; as shown in Figure~\ref{fig:block_length_comparison}, $S^3$ recovers its advantage at finer block lengths ($\mathcal{K} \in \{2,4,16\}$), where the look-ahead signal is more informative. Crucially, $S^3$ improves over BoK on MATH-500, GSM8K, and TruthfulQA, confirming that the gains arise from reallocating compute \emph{during} denoising rather than merely increasing the final sample count.

\subsection{Scaling with Inference-Time Compute and Block-Length Analysis}
Figure~\ref{fig:multi_dataset_scaling} (top row) shows accuracy vs.\ NFE across multiple $(N,b)$ configurations. $S^3$ surpasses the BoK Pareto frontier at matched NFE on MATH-500, GSM8K, and TruthfulQA; on ARC-Challenge, $S^3$ remains competitive at low NFE but does not dominate at higher budgets, consistent with the weaker verifier signal on multiple-choice tasks. The confidence evolution plots (bottom row) show that $S^3$ maintains higher mean token confidence throughout denoising compared to the baseline, indicating that verifier-guided resampling stabilizes intermediate generation dynamics rather than only affecting the final output selection. Figure~\ref{fig:block_length_comparison} shows that $S^3$ consistently outperforms both baselines across all block lengths on MATH-500 and GSM8K, and matches or exceeds BoK on TruthfulQA. On ARC-Challenge, $S^3$ leads at block lengths $\mathcal{K} \in \{2, 4, 16\}$ but underperforms BoK at $\mathcal{K} \in \{8, 32, 64\}$, suggesting that the look-ahead signal is less reliable at coarser granularities on multiple-choice tasks where the verifier provides weaker per-step discrimination. Full latency and output token statistics across all block sizes are reported in Appendix~\ref{app:ablation_cost}.

\subsection{Ablation: Tilted Reward vs.\ Look-ahead} \label{sec:ablation}
To isolate the contribution of each component, we evaluate four conditions under a matched compute budget of $K = N \cdot b$ forward passes: (1) \textbf{Baseline diffusion}, single-trajectory decoding with no verifier; (2) \textbf{Best-of-$K$}, $K$ independent sequences from $p_0$ with majority voting and NLL tie-breaking; (3) \textbf{Look-ahead only}, tree expansion with uniform top-$N$ pruning instead of exponential weighting, with majority voting and NLL tie-breaking over surviving particles; and (4) \textbf{Tilting only}, $K$ independent sequences reweighted by $w_i \propto \exp(\lambda f(x_i))$ without tree structure, selected via weighted argmax. The full $S^3$ method uses both look-ahead and tilting with majority voting and NLL tie-breaking. Table~\ref{tab:ablation_n4b2} shows that neither look-ahead nor tilting alone is sufficient in isolation: look-ahead without tilting yields marginal gains, and tilting without look-ahead can underperform BoK on MATH-500. Only their combination in $S^3$ achieves consistent improvements, with the largest synergy on MATH-500 (+2.40 pp over BoK), empirically validating the two-component design motivated by the approximate twisted SMC framework in Section~\ref{sec:method}.

\begin{table}[t]
\centering
\small
\caption{%
  Ablation study decomposing the contributions of look-ahead search and
  Gibbs tilting in $S^3$ (LLaDA-8B-Instruct, $N{=}4$, $b{=}2$,
  $K{=}N{\cdot}b{=}8$, block length 64, generation length 128).
  \checkmark/\texttimes{} indicate whether each mechanism is active.
  Scores represent task accuracy (\%).%
}
\label{tab:ablation_n4b2}
\setlength{\tabcolsep}{4pt}
\begin{tabular}{lcccccc}
\toprule
\multirow{2}{*}{\textbf{Method}}
  & \multirow{2}{*}{\textbf{Look-ahead}}
  & \multirow{2}{*}{\textbf{Tilting}}
  & \multicolumn{4}{c}{\textbf{Accuracy (\%)}} \\
\cmidrule(lr){4-7}
  & & & \textbf{GSM8K} & \textbf{MATH-500} & \textbf{ARC-C} & \textbf{TruthfulQA} \\
\midrule
Baseline diffusion
  & $\times$ & $\times$
  & 68.16 & 25.60 & 76.11 & 46.49 \\
Best-of-$K$
  & $\times$ & $\times$
  & \underline{69.56} & \underline{28.20} & \textbf{79.30} & \underline{49.36} \\
Tilting only
  & $\times$ & \checkmark
  & 69.54 & 26.40 & 76.30 & 48.10 \\
Look-ahead only
  & \checkmark & $\times$
  & 69.69 & 26.20 & 76.71 & 46.49 \\
$S^3$ (full, ours)
  & \checkmark & \checkmark
  & \textbf{70.21} & \textbf{30.20} & \underline{77.86} & \textbf{49.57} \\
\bottomrule
\end{tabular}
\end{table}

\section{Related Work} \label{sec:related}
\paragraph{Diffusion models for generation.}
Score-based diffusion models learn to reverse a noising process to generate samples \citep{song2020score}, and have been extended to discrete domains via masked denoising \citep{austin2021structured}. In NLP, Diffusion-LM \citep{li2022diffusion} introduced diffusion-based text generation, with subsequent work scaling DLMs to competitive performance \citep{nie2025large}.

\paragraph{Inference-time control and test-time scaling.}
Recent studies show that allocating additional inference-time compute can substantially improve model performance \citep{snell2024scaling}. This has motivated research on verifier design and compute allocation strategies \citep{chen2025rethinking}. Exponential tilting of base distributions toward reward-weighted targets provides a principled foundation for inference-time objective formulations \citep{donsker1975asymptotic}. For diffusion models, several approaches steer generation at inference time without retraining, including training-free guidance \citep{ye2024tfg}, derivative-free or value-based decoding \citep{li2024derivative}, and controlled Langevin diffusion processes \citep{chen2024sequential}. Analogous verifier-guided scaling has also been explored for vision diffusion models \citep{ma2025scaling, baraldi2025verifier}. These methods modify the denoising dynamics to incorporate external reward signals during sampling; however, their reliance on continuous state spaces and gradient-based updates renders them incompatible with discrete masked DLMs, and thus not directly comparable to $S^3$.

\paragraph{Population and search-based diffusion inference.}
Another line of work scales diffusion inference via population-based methods that maintain and resample multiple denoising trajectories \citep{dang2025inference}, building on SMC \citep{smith2013sequential} and annealed importance sampling \citep{neal2001annealed} as general frameworks for importance-weighted resampling over sequences of target distributions. The auxiliary particle filter \citep{pitt1999filtering} and twisted particle filters \citep{whiteley2014twisted} provide the theoretical foundation for $S^3$ via look-ahead weighting and backward information function approximation, respectively. Orthogonal directions include amortizing inference-time compute via auxiliary networks \citep{eyring2025noise} and scaling compute for vision diffusion models \citep{ma2025scaling, baraldi2025verifier}. Closely related work interprets diffusion decoding as search over denoising trajectories and applies classical search strategies such as BFS or DFS to guide exploration \citep{search_diffusion}, and hierarchical search with self-verification has been explored for discrete DLMs \citep{bai2026prism}. Both are complementary to $S^3$ rather than directly comparable, as neither grounds search in a principled inference-time objective nor employs verifier-weighted SMC resampling.

\section{Conclusion} \label{sec:conclusion}
This work demonstrates that test-time scaling for DLMs improves by reallocating compute across denoising trajectories rather than increasing the final sample count. $S^3$ performs verifier-guided particle search that shifts generation toward higher-quality output regions without modifying the underlying model or decoding schedule, yielding consistent gains across four benchmarks. The primary limitation is the reliance on verifier quality and intermediate clean prediction accuracy: noisy or misaligned signals may fail to guide trajectories toward better solutions. The method also incurs additional compute due to particle expansion and repeated scoring, making the compute-performance trade-off an important consideration for practical deployment.

\section*{Ethics Statement}
This work studies inference-time compute allocation for DLMs and does not introduce new datasets, human subjects, or deployment systems that raise additional ethical concerns beyond those associated with large language models.

\section*{Reproducibility Statement}
All experiments use publicly available benchmarks and a fixed model configuration, and we provide the evaluation setup, algorithm description, and hyperparameters necessary to reproduce the results.

\section*{Usage of LLMs}
We used LLMs to assist with the research, coding, and writing of this paper, in accordance with COLM policy.

LLMs were used to refine and streamline text, support aspects of the literature review, and assist with reference formatting. All content has been carefully reviewed, and we take full responsibility for the accuracy and validity of the paper.

We also used LLMs as coding assistants for implementation, validation, and plotting. The resulting code has been thoroughly tested, sanity-checked, and manually reviewed. We stand by the correctness of the software and the claims derived from it.

\section*{Acknowledgments}
We thank the anonymous reviewers and the open-source community whose tools and datasets made this research possible.

\bibliography{main}

@article{search_diffusion,
  title={Inference-time scaling of diffusion models through classical search},
  author={Zhang, Xiangcheng and Lin, Haowei and Ye, Haotian and Zou, James and Ma, Jianzhu and Liang, Yitao and Du, Yilun},
  journal={arXiv preprint arXiv:2505.23614},
  year={2025}
}

@article{dang2025inference,
  title={Inference-time scaling of diffusion language models with particle gibbs sampling},
  author={Dang, Meihua and Han, Jiaqi and Xu, Minkai and Xu, Kai and Srivastava, Akash and Ermon, Stefano},
  journal={arXiv preprint arXiv:2507.08390},
  year={2025}
}

@inproceedings{ma2025scaling,
  title={Scaling inference time compute for diffusion models},
  author={Ma, Nanye and Tong, Shangyuan and Jia, Haolin and Hu, Hexiang and Su, Yu-Chuan and Zhang, Mingda and Yang, Xuan and Li, Yandong and Jaakkola, Tommi and Jia, Xuhui and others},
  booktitle={Proceedings of the Computer Vision and Pattern Recognition Conference},
  pages={2523--2534},
  year={2025}
}

@inproceedings{baraldi2025verifier,
  title={Verifier Matters: Enhancing Inference-Time Scaling for Video Diffusion Models},
  author={Baraldi, Lorenzo and Bucciarelli, Davide and Zeng, Zifan and Zhang, Chongzhe and Zhang, Qunli and Cornia, Marcella and Liu, Feng and Zheng, Hu and Cucchiara, Rita and others},
  booktitle={Proceedings of the 36th British Machine Vision Conference},
  year={2025}
}

@article{song2020score,
  title={Score-based generative modeling through stochastic differential equations},
  author={Song, Yang and Sohl-Dickstein, Jascha and Kingma, Diederik P and Kumar, Abhishek and Ermon, Stefano and Poole, Ben},
  journal={arXiv preprint arXiv:2011.13456},
  year={2020}
}

@article{chen2024sequential,
  title={Sequential controlled langevin diffusions},
  author={Chen, Junhua and Richter, Lorenz and Berner, Julius and Blessing, Denis and Neumann, Gerhard and Anandkumar, Anima},
  journal={arXiv preprint arXiv:2412.07081},
  year={2024}
}

@article{ye2024tfg,
  title={Tfg: Unified training-free guidance for diffusion models},
  author={Ye, Haotian and Lin, Haowei and Han, Jiaqi and Xu, Minkai and Liu, Sheng and Liang, Yitao and Ma, Jianzhu and Zou, James Y and Ermon, Stefano},
  journal={Advances in Neural Information Processing Systems},
  volume={37},
  pages={22370--22417},
  year={2024}
}

@article{austin2021structured,
  title={Structured denoising diffusion models in discrete state-spaces},
  author={Austin, Jacob and Johnson, Daniel D and Ho, Jonathan and Tarlow, Daniel and Van Den Berg, Rianne},
  journal={Advances in neural information processing systems},
  volume={34},
  pages={17981--17993},
  year={2021}
}

@article{bai2026prism,
  title={Prism: Efficient Test-Time Scaling via Hierarchical Search and Self-Verification for Discrete Diffusion Language Models},
  author={Bai, Jinbin and Li, Yixuan and Zhu, Yuchen and Xin, Yi and Shi, Qingyu and Feng, Aosong and Liu, Xiaohong and Tao, Molei and Xue, Jianru and Li, Xiangtai and others},
  journal={arXiv preprint arXiv:2602.01842},
  year={2026}
}

@article{li2022diffusion,
  title={Diffusion-lm improves controllable text generation},
  author={Li, Xiang and Thickstun, John and Gulrajani, Ishaan and Liang, Percy S and Hashimoto, Tatsunori B},
  journal={Advances in neural information processing systems},
  volume={35},
  pages={4328--4343},
  year={2022}
}

@article{li2024derivative,
  title={Derivative-free guidance in continuous and discrete diffusion models with soft value-based decoding},
  author={Li, Xiner and Zhao, Yulai and Wang, Chenyu and Scalia, Gabriele and Eraslan, Gokcen and Nair, Surag and Biancalani, Tommaso and Ji, Shuiwang and Regev, Aviv and Levine, Sergey and others},
  journal={arXiv preprint arXiv:2408.08252},
  year={2024}
}

@article{snell2024scaling,
  title={Scaling llm test-time compute optimally can be more effective than scaling model parameters},
  author={Snell, Charlie and Lee, Jaehoon and Xu, Kelvin and Kumar, Aviral},
  journal={arXiv preprint arXiv:2408.03314},
  year={2024}
}

@article{chen2025rethinking,
  title={Rethinking Optimal Verification Granularity for Compute-Efficient Test-Time Scaling},
  author={Chen, Hao Mark and Lu, Guanxi and Okoshi, Yasuyuki and Mo, Zhiwen and Motomura, Masato and Fan, Hongxiang},
  journal={arXiv preprint arXiv:2505.11730},
  year={2025}
}

@article{eyring2025noise,
  title={Noise hypernetworks: Amortizing test-time compute in diffusion models},
  author={Eyring, Luca and Karthik, Shyamgopal and Dosovitskiy, Alexey and Ruiz, Nataniel and Akata, Zeynep},
  journal={arXiv preprint arXiv:2508.09968},
  year={2025}
}

@article{brown2024largelanguagemonkeysscaling,
  title={Large language monkeys: Scaling inference compute with repeated sampling},
  author={Brown, Bradley and Juravsky, Jordan and Ehrlich, Ryan and Clark, Ronald and Le, Quoc V and R{\'e}, Christopher and Mirhoseini, Azalia},
  journal={arXiv preprint arXiv:2407.21787},
  year={2024}
}

@article{zhao2025d1,
  title={d1: Scaling reasoning in diffusion large language models via reinforcement learning},
  author={Zhao, Siyan and Gupta, Devaansh and Zheng, Qinqing and Grover, Aditya},
  journal={arXiv preprint arXiv:2504.12216},
  year={2025}
}

@article{hendrycks2021measuring,
  title={Measuring mathematical problem solving with the math dataset},
  author={Hendrycks, Dan and Burns, Collin and Kadavath, Saurav and Arora, Akul and Basart, Steven and Tang, Eric and Song, Dawn and Steinhardt, Jacob},
  journal={arXiv preprint arXiv:2103.03874},
  year={2021}
}

@book{concentration_inequality,
    author = {Boucheron, Stéphane and Lugosi, Gábor and Massart, Pascal},
    title = {Concentration Inequalities: A Nonasymptotic Theory of Independence},
    publisher = {Oxford University Press},
    year = {2013},
    month = {02},
    isbn = {9780199535255},
    doi = {10.1093/acprof:oso/9780199535255.001.0001},
    url = {https://doi.org/10.1093/acprof:oso/9780199535255.001.0001},
}

@article{donsker1975asymptotic,
  title={Asymptotic evaluation of certain Markov process expectations for large time, I},
  author={Donsker, Monroe D and Varadhan, SR Srinivasa},
  journal={Communications on pure and applied mathematics},
  volume={28},
  number={1},
  pages={1--47},
  year={1975},
  publisher={Wiley Online Library}
}

@article{cobbe2021training,
  title={Training verifiers to solve math word problems},
  author={Cobbe, Karl and Kosaraju, Vineet and Bavarian, Mohammad and Chen, Mark and Jun, Heewoo and Kaiser, Lukasz and Plappert, Matthias and Tworek, Jerry and Hilton, Jacob and Nakano, Reiichiro and others},
  journal={arXiv preprint arXiv:2110.14168},
  year={2021}
}

@article{clark2018think,
  title={Think you have solved question answering? try arc, the ai2 reasoning challenge},
  author={Clark, Peter and Cowhey, Isaac and Etzioni, Oren and Khot, Tushar and Sabharwal, Ashish and Schoenick, Carissa and Tafjord, Oyvind},
  journal={arXiv preprint arXiv:1803.05457},
  year={2018}
}

@inproceedings{lin2022truthfulqa,
  title={Truthfulqa: Measuring how models mimic human falsehoods},
  author={Lin, Stephanie and Hilton, Jacob and Evans, Owain},
  booktitle={Proceedings of the 60th annual meeting of the association for computational linguistics (volume 1: long papers)},
  pages={3214--3252},
  year={2022}
}

@article{nie2025large,
  title={Large language diffusion models},
  author={Nie, Shen and Zhu, Fengqi and You, Zebin and Zhang, Xiaolu and Ou, Jingyang and Hu, Jun and Zhou, Jun and Lin, Yankai and Wen, Ji-Rong and Li, Chongxuan},
  journal={arXiv preprint arXiv:2502.09992},
  year={2025}
}

@article{srinivasan2001distributions,
  title={Improved approximation guarantees for packing and covering integer programs},
  author={Srinivasan, Aravind},
  journal={SIAM Journal on Computing},
  volume={29},
  number={2},
  pages={648--670},
  year={1999},
  publisher={SIAM}
}

@article{kim2025train,
  title={Train for the worst, plan for the best: Understanding token ordering in masked diffusions},
  author={Kim, Jaeyeon and Shah, Kulin and Kontonis, Vasilis and Kakade, Sham and Chen, Sitan},
  journal={arXiv preprint arXiv:2502.06768},
  year={2025}
}

@article{nie2024scaling,
  title={Scaling up masked diffusion models on text},
  author={Nie, Shen and Zhu, Fengqi and Du, Chao and Pang, Tianyu and Liu, Qian and Zeng, Guangtao and Lin, Min and Li, Chongxuan},
  journal={arXiv preprint arXiv:2410.18514},
  year={2024}
}

@article{lee2025test,
  title={Test-Time Scaling in Diffusion LLMs via Hidden Semi-Autoregressive Experts},
  author={Lee, Jihoon and Moon, Hoyeon and Zhai, Kevin and Chithanar, Arun Kumar and Sahu, Anit Kumar and Kar, Soummya and Lee, Chul and Chakraborty, Souradip and Bedi, Amrit Singh},
  journal={arXiv preprint arXiv:2510.05040},
  year={2025}
}

@inproceedings{do2025discrete,
  title={Discrete diffusion language model for efficient text summarization},
  author={Do, Duc Anh and Tuan, Luu Anh and Buntine, Wray and others},
  booktitle={Findings of the Association for Computational Linguistics: NAACL 2025},
  pages={6278--6290},
  year={2025}
}

@article{xu2024energy,
  title={Energy-based diffusion language models for text generation},
  author={Xu, Minkai and Geffner, Tomas and Kreis, Karsten and Nie, Weili and Xu, Yilun and Leskovec, Jure and Ermon, Stefano and Vahdat, Arash},
  journal={arXiv preprint arXiv:2410.21357},
  year={2024}
}

@book{smith2013sequential,
  title={Sequential Monte Carlo methods in practice},
  author={Smith, Adrian},
  year={2013},
  publisher={Springer Science \& Business Media}
}

@article{whiteley2014twisted,
  title={Twisted particle filters},
  author={Whiteley, Nick and Lee, Anthony},
  year={2014}
}

@article{neal2001annealed,
  title={Annealed importance sampling},
  author={Neal, Radford M},
  journal={Statistics and computing},
  volume={11},
  number={2},
  pages={125--139},
  year={2001},
  publisher={Springer}
}

@article{pitt1999filtering,
  title={Filtering via simulation: Auxiliary particle filters},
  author={Pitt, Michael K and Shephard, Neil},
  journal={Journal of the American statistical association},
  volume={94},
  number={446},
  pages={590--599},
  year={1999},
  publisher={Taylor \& Francis}
}
\bibliographystyle{colm2026_conference}

\appendix
\section{Distributional Shift of Verifier Scores}
\label{app:dist_shift}

\begin{figure}[h]
  \centering
  \begin{subfigure}[t]{0.48\textwidth}
    \centering
    \includegraphics[width=\textwidth]{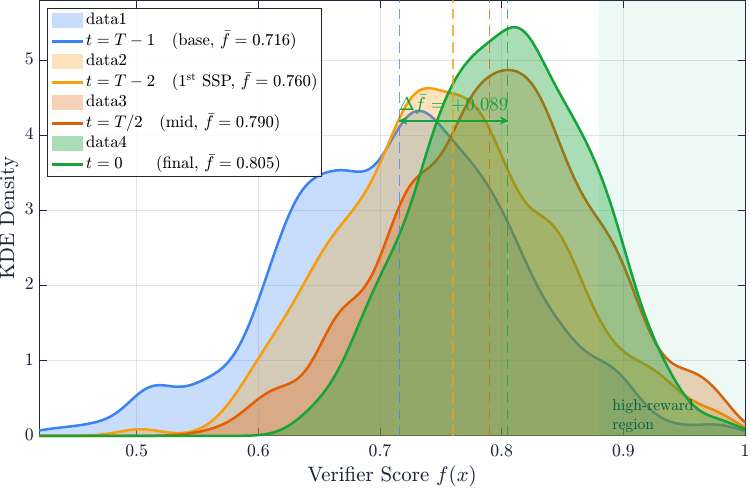}
    \caption{KDE of verifier scores across denoising steps.}
    \label{fig:dist_shift_kde}
  \end{subfigure}
  \hfill
  \begin{subfigure}[t]{0.48\textwidth}
    \centering
    \includegraphics[width=\textwidth]{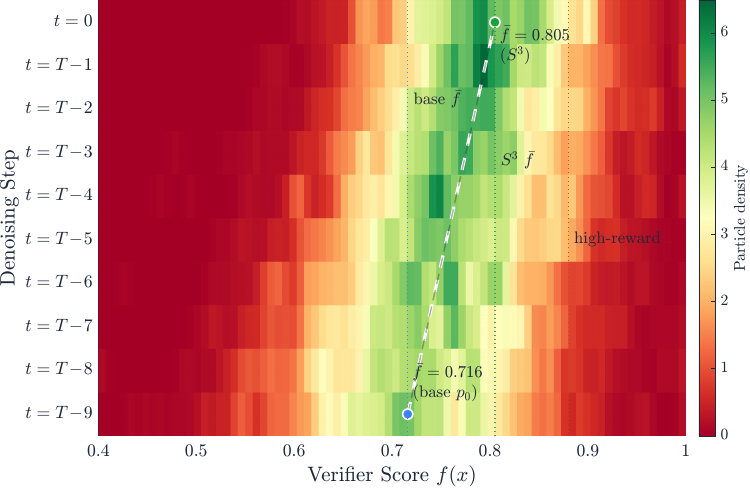}
    \caption{Particle score density over denoising steps.}
    \label{fig:dist_shift_heatmap}
  \end{subfigure}
\caption{Distributional shift of verifier scores under $S^3$ search on MATH-500. Sequential resampling during denoising concentrates particles in higher-reward regions compared to direct sampling from $p_0$.}
\label{fig:dist_shift}
\end{figure}

Figure~\ref{fig:dist_shift} illustrates the distributional shift of verifier scores across denoising steps under $S^3$ search on MATH-500. Sequential resampling progressively concentrates particles in higher-reward regions compared to direct sampling from $p_0$. The KDE plot (Figure~\ref{fig:dist_shift_kde}) shows the score distribution shifting rightward across steps, while the heatmap (Figure~\ref{fig:dist_shift_heatmap}) shows the particle score density concentrating toward higher-reward regions as denoising proceeds, confirming that the expand--score--resample updates progressively move the particle population away from the unguided base path measure $p$ and toward the reward-tilted distribution in Eq.~\eqref{eq:tilted_path}.

\section{Experimental Details}
\label{app:exp_details}

All experiments use \texttt{LLaDA-8B-Instruct} with \texttt{gen\_length}=128 and \texttt{diffusion\_steps}=64. We follow the evaluation protocol of \citet{lee2025test} for dataset preprocessing and answer normalization: numerical and symbolic answers are extracted via \LaTeX{} canonicalization for GSM8K and MATH-500, and multiple-choice predictions are mapped to option sets for ARC-Challenge and TruthfulQA. GSM8K comprises 1,319 test problems, MATH-500 comprises 500 competition-level problems, TruthfulQA comprises 817 questions, and ARC-Challenge comprises 1,172 questions. The $S^3$ default configuration uses $N{=}4$, $b{=}2$, $\lambda{=}1.0$, and $\mathcal{K}{=}64$ unless otherwise stated.

\paragraph{Benchmark Performance.}
On MATH-500, $S^3$ improves accuracy from 25.60\% to 30.20\% (+4.60 pp over baseline, +2.00 pp over BoK). On GSM8K, accuracy improves from 68.16\% to 70.21\% (+2.05 pp over baseline, +0.65 pp over BoK). On TruthfulQA, accuracy improves from 46.49\% to 49.57\% (+3.08 pp over baseline, +0.21 pp over BoK). On ARC-Challenge, accuracy improves from 76.11\% to 77.86\% (+1.75 pp over baseline, $-$1.44 pp vs.\ best-of-$K$), with $S^3$ recovering its advantage over best-of-$K$ at finer block lengths as discussed in Section~\ref{sec:eval}.
\section{Composite Verifier Design}
\label{app:verifier}

The verifier $f : \mathcal{X}^L \to [0,1]$ must satisfy two operational 
requirements: it must produce a meaningful quality signal \emph{without access 
to ground-truth answers}, and it must be \emph{inexpensive enough to evaluate 
$Nb$ times per denoising step} throughout the $S^3$ search procedure. To satisfy 
both constraints simultaneously, we design a lightweight \textbf{composite 
intrinsic verifier} that decomposes solution quality into five orthogonal 
dimensions derivable from the generated text alone, plus an optional 
task-specific constraint term.

\subsection{Composite Score Formulation}

The verifier score is defined as
\begin{equation}
f(x) \;=\; \sum_{k=1}^{5} \alpha_k\, s_k(x) \;+\; \alpha_c\, 
s_{\mathrm{constraint}}(x),
\label{eq:verifier}
\end{equation}
where $s_1,\dots,s_5$ measure structural completeness, internal arithmetic 
consistency, answer reachability, model confidence, and non-degeneracy 
respectively; $s_{\mathrm{constraint}}$ captures task-specific constraint 
satisfaction, and the coefficients satisfy
\[
\sum_{k=1}^{5} \alpha_k + \alpha_c = 1, \qquad \alpha_k \ge 0.
\]
Each component is described in detail below. The five dimensions are 
intentionally orthogonal: structural completeness captures surface form, 
arithmetic consistency captures local numerical correctness, answer reachability 
captures global reasoning coherence, model confidence captures fluency, and 
non-degeneracy penalizes degenerate failure modes. No single component is 
sufficient alone; the composite aggregation makes the signal robust to the 
weaknesses of any individual dimension.

\subsection{Dataset-Specific Weight Profiles} \label{app:weights}
The coefficients $(\alpha_1,\dots,\alpha_5,\alpha_c)$ are heuristically set based on the expected answer format: numerical answer extraction for mathematical reasoning tasks (GSM8K, MATH-500) and letter option selection for multiple-choice tasks (ARC-Challenge, TruthfulQA). Consequently, arithmetic consistency $\alpha_2$ and answer reachability $\alpha_3$ carry the largest weight for mathematical tasks, while they are zeroed out for multiple-choice tasks where the constraint term $\alpha_c$ dominates, as summarized in Table~\ref{tab:weight_profiles}.

\begin{table}[h]
\centering
\small
\caption{Weight profiles $(\alpha_1,\dots,\alpha_5,\alpha_c)$ per dataset, 
corresponding to structural completeness, arithmetic consistency, and answer 
reachability, model confidence, non-degeneracy, and constraint satisfaction 
respectively. All rows sum to $1.0$.}
\label{tab:weight_profiles}
\setlength{\tabcolsep}{6pt}
\begin{tabular}{lcccccc}
\toprule
\textbf{Dataset} 
  & $\alpha_1$ & $\alpha_2$ & $\alpha_3$ 
  & $\alpha_4$ & $\alpha_5$ & $\alpha_c$ \\
\midrule
GSM8K      & 0.20 & 0.25 & 0.25 & 0.10 & 0.20 & 0.00 \\
MATH-500   & 0.25 & 0.20 & 0.25 & 0.10 & 0.20 & 0.00 \\
ARC-Challenge      & 0.20 & 0.00 & 0.00 & 0.15 & 0.20 & 0.45 \\
TruthfulQA & 0.20 & 0.00 & 0.00 & 0.15 & 0.20 & 0.45 \\
Countdown  & 0.15 & 0.10 & 0.10 & 0.10 & 0.15 & 0.40 \\
Sudoku     & 0.05 & 0.00 & 0.00 & 0.05 & 0.10 & 0.80 \\
\bottomrule
\end{tabular}
\end{table}

For \textbf{mathematical reasoning} (GSM8K, MATH-500), arithmetic consistency 
$\alpha_2$ and answer reachability $\alpha_3$ together carry the largest combined 
weight ($0.50$ for GSM8K; $0.45$ for MATH-500). MATH-500 upweights structural 
completeness ($\alpha_1 = 0.25$) relative to GSM8K ($\alpha_1 = 0.20$) to 
account for the more elaborate solution formats typical of competition mathematics, 
while GSM8K slightly upweights arithmetic consistency ($\alpha_2 = 0.25$) to 
capture the denser arithmetic chains present in grade-school word problems. The 
constraint term is set to zero for both, since no domain constraints are 
verifiable without ground truth.

For \textbf{multiple-choice tasks} (ARC-Challenge, TruthfulQA), the arithmetic 
components $\alpha_2$ and $\alpha_3$ are zeroed out, as these benchmarks require 
answer selection rather than numerical derivation. The constraint term carries 
$\alpha_c = 0.45$ via an MC reasoning quality score that rewards the answer 
extraction, reasoning keyword presence, answer dominance across options, and 
minimum response length.

For \textbf{structured constraint tasks}, the constraint term dominates: $\alpha_c = 0.40$ for Countdown (target match and number validity) and $\alpha_c = 0.80$ for Sudoku (grid validity and clue preservation), since these tasks are self-verifying, and domain constraint satisfaction is a far stronger quality signal than any text-surface heuristic.

\subsection{Component Scores}

\paragraph{S1: Structural Completeness.}
$s_{\mathrm{struct}}(x)$ rewards outputs that exhibit the surface form of 
well-structured solutions. For mathematical reasoning tasks, the score aggregates 
three sub-signals: (i) a soft-thresholded keyword match $\min(h/3, 1.0)$ over 
a vocabulary of 14 reasoning keywords (e.g., \emph{step}, \emph{therefore}, 
\emph{thus}, \emph{compute}), where $h$ is the number of distinct keyword hits; 
(ii) a binary answer-delimiter check using the priority cascade 
$\verb|\boxed{...}| \succ \verb|<answer>...</answer>| \succ 
\texttt{\#\#\#\#} \succ \text{``answer is/=/''}$; and (iii) an alphanumeric 
density score $\min(r/0.5, 1.0)$ where $r$ is the fraction of alphanumeric 
or whitespace characters. For multiple-choice tasks, the score reduces to a 
binary extraction check for a letter option in $\{\mathrm{A},\dots,\mathrm{H}\}$ 
plus a keyword match and density term.

\begin{tcolorbox}[colback=gray!5!white, colframe=black!40, boxrule=0.6pt,
  arc=3pt, left=5pt, right=5pt, top=3pt, bottom=3pt]
\small
\textbf{Example.} For the problem \emph{``If $3x+5=14$, find $x$''}, the output
\emph{``Step 1: subtract 5: $3x=9$. Therefore $x=3$. $\backslash$boxed\{3\}''}
receives $s_{\mathrm{struct}}=1.0$ (keywords present, boxed answer, high 
alphanumeric density). The degenerate output \emph{``3''} receives 
$s_{\mathrm{struct}}=0.25$ (answer present but no reasoning structure).
\end{tcolorbox}

\textit{Limitation.} Keyword matching is purely lexical. A response of the form 
``step step therefore $\backslash$boxed\{42\}'' scores highly on structure 
despite being semantically vacuous.

\paragraph{S2: Internal Arithmetic Consistency.}
$s_{\mathrm{consist}}(x)$ verifies explicit arithmetic equalities of the form 
$a \;\mathrm{op}\; b = c$ for $\mathrm{op} \in \{+,-,\times,\div\}$ appearing in 
the generated reasoning trace, in both ASCII and \LaTeX{} notation. A tolerance 
of $\max(|\hat{c}| \times 10^{-6},\, 10^{-4})$ is applied to the stated result 
$\hat{c}$ to handle floating-point rounding. An additional chain-consistency 
check verifies that consecutive equality values 
$\{v_1, v_2, \dots\}$ extracted via the pattern $= \langle\text{number}\rangle$ 
have consecutive ratios within $(10^{-3}, 10^3)$, penalizing outputs whose 
intermediate values diverge implausibly. The score is
\[
s_{\mathrm{consist}}(x) \;=\; 
\begin{cases}
\dfrac{|\{\text{verified equalities}\}|}{|\{\text{total equalities}\}|} 
  & \text{if any equalities detected,} \\[6pt]
0.5 & \text{if no equalities detected (neutral default).}
\end{cases}
\]

\begin{tcolorbox}[colback=gray!5!white, colframe=black!40, boxrule=0.6pt,
  arc=3pt, left=5pt, right=5pt, top=3pt, bottom=3pt]
\small
\textbf{Example.} The trace \emph{``$12 \times 4 = 48$, then $48 + 7 = 55$''} 
yields $s_{\mathrm{consist}} = 1.0$ (both equalities verified). The trace 
\emph{``$12 \times 4 = 50$, then $50 + 7 = 55$''} yields 
$s_{\mathrm{consist}} = 0.5$ (one of two equalities incorrect). A response 
containing no arithmetic expressions yields $s_{\mathrm{consist}} = 0.5$ 
(neutral default, not penalized).
\end{tcolorbox}

\textit{Limitation.} The parser does not handle compound parenthesized 
expressions (e.g., $(a+b)\times c = d$), multi-line equations, or symbolic 
variables. Logical errors not expressed as explicit equalities (e.g., incorrect 
substitutions) are invisible to this component.

\paragraph{S3: Answer Reachability.}
$s_{\mathrm{reach}}(x)$ checks whether the final answer can be traced back to 
the preceding reasoning. The final answer is extracted via the same priority 
cascade as $s_{\mathrm{struct}}$. The reasoning prefix is defined as all text 
before the detected answer delimiter. For numeric answers, the check verifies 
the presence of the same floating-point value (to tolerance $10^{-6}$) among 
all numbers in the reasoning prefix; for symbolic answers, case-insensitive 
substring matching is used. The score is
\[
s_{\mathrm{reach}}(x) \;=\; 
\begin{cases}
1.0 & \text{answer found and traceable to reasoning,} \\
0.3 & \text{answer found but not traceable,} \\
0.2 & \text{no answer found.}
\end{cases}
\]

\begin{tcolorbox}[colback=gray!5!white, colframe=black!40, boxrule=0.6pt,
  arc=3pt, left=5pt, right=5pt, top=3pt, bottom=3pt]
\small
\textbf{Example.} The output \emph{``We compute $6\times 7=42$. The answer is 
$\backslash$boxed\{42\}''} scores $s_{\mathrm{reach}}=1.0$. The output 
\emph{``$\backslash$boxed\{42\}''} with no preceding reasoning scores 
$s_{\mathrm{reach}}=0.3$ (answer present but unreachable from the empty prefix).
\end{tcolorbox}

\textit{Limitation.} Reachability is checked by value occurrence, not logical 
derivation. A response that mentions the correct value incidentally 
(e.g., ``this is not 42'') would still receive full reachability credit.

\paragraph{S4: Model Confidence.}
When token-level log-probabilities are available, we compute
\[
s_{\mathrm{conf}}(x) \;=\;
\frac{1}{|G|}\sum_{i \in G} p_\theta(x_i \mid x_{<i}),
\]
where $G$ denotes the generated token positions (excluding the prompt). 
Logits are clamped to $[-100, 100]$ before softmax to prevent numerical 
overflow, and the result is clipped to $[0,1]$. This serves as a proxy for 
local fluency and coherence. When logits are unavailable (e.g., black-box 
APIs), the score defaults to $0.5$ to avoid biasing the composite.

\textit{Limitation.} Model confidence reflects the model's own distribution 
and not solution correctness. High-confidence outputs can still be factually 
wrong, particularly when the model is confidently biased toward plausible but 
incorrect completions. This component is therefore most informative in 
combination with the other four signals rather than in isolation.

\paragraph{S5: Non-degeneracy.}
$s_{\mathrm{ndegen}}(x)$ penalizes collapsed or low-information outputs via 
a cascade of hard and soft thresholds:
\begin{enumerate}[leftmargin=1.5em,itemsep=2pt]
\item \textit{Special token flood}: if \texttt{<|endoftext|>} or 
\texttt{<|eot\_id|>} tokens exceed 20\% of word positions, return $0.0$.
\item \textit{Unresolved mask tokens}: if \texttt{<|mdm\_mask|>} tokens exceed 
15\% of positions return $0.05$.
\item \textit{Insufficient length}: fewer than 8 words returns $0.2$.
\item \textit{Bigram diversity} (outputs $>12$ words): unique bigram ratio 
$\rho_2 = |\text{unique bigrams}|/|\text{total bigrams}|$; return $0.05$ if 
$\rho_2 < 0.15$, return $0.3$ if $\rho_2 < 0.30$.
\item \textit{Trigram diversity} (outputs $>30$ words): unique trigram ratio 
$\rho_3 < 0.25$ returns $0.2$.
\item \textit{Otherwise}: return $1.0$.
\end{enumerate}

\begin{tcolorbox}[colback=gray!5!white, colframe=black!40, boxrule=0.6pt,
  arc=3pt, left=5pt, right=5pt, top=3pt, bottom=3pt]
\small
\textbf{Example.} The output \emph{``the answer is 42 the answer is 42 the 
answer is 42''} (11 bigrams, 2 unique) yields $\rho_2 \approx 0.18 < 0.30$, 
returning $s_{\mathrm{ndegen}} = 0.3$. A fully collapsed output of repeated 
\texttt{<|endoftext|>} tokens returns $s_{\mathrm{ndegen}} = 0.0$ immediately.
\end{tcolorbox}

\textit{Limitation.} Very short outputs (fewer than 8 words) bypass the 
repetition checks and are penalized only by the length threshold. Outputs 
consisting of diverse but semantically vacuous tokens score 
$s_{\mathrm{ndegen}} = 1.0$ despite being uninformative.

\subsection{Task-Specific Constraint Term}
\label{app:constraints}

\paragraph{Countdown.}
The constraint score $s_{\mathrm{constraint}}$ combines a target-match sub-score 
(weight $0.6$) and a number-validity sub-score (weight $0.4$). Both the target 
value and the allowed number pool are parsed \emph{solely from} 
\texttt{input\_text} via pattern matching (e.g., ``Target: 952 Numbers: 100, 75, 
50, 25, 6, 3''), with no ground-truth metadata passed in. The target-match 
sub-score is $0.6$ if the extracted arithmetic expression evaluates to within 
$10^{-6}$ of the target, $0.3$ if within $\pm 1$, and $0.1$ otherwise. The 
number-validity sub-score is $0.4$ if all operands are drawn from the allowed 
pool with correct multiplicity, and $0.1$ otherwise. If neither target nor 
allowed numbers can be parsed from \texttt{input\_text}, the score falls back to 
$0.7$ when any evaluable arithmetic expression is present, and $0.0$ otherwise.

\paragraph{Sudoku.}
Sudoku is self-verifying: a solution is valid if and only if all 27 
row/column/box uniqueness constraints are satisfied. The constraint score is
\[
s_{\mathrm{constraint}}(x) 
\;=\; 0.75 \cdot \frac{|\text{satisfied constraints}|}{27} 
\;+\; 0.25 \cdot s_{\mathrm{clue}}(x),
\]
where $s_{\mathrm{clue}} \in [0,1]$ measures the fraction of given puzzle clues 
(parsed from \texttt{input\_text} as an 81-character digit string) that are 
correctly preserved in the output. This score is fully ground-truth-free: 
correctness of a completed Sudoku grid is entirely determined by the structural 
constraints of the puzzle itself.

\paragraph{Multiple-choice (ARC-Challenge, TruthfulQA).}
The constraint term is replaced by an MC reasoning quality score with four 
additive components: (i) extraction of a clear letter choice from 
$\{\mathrm{A},\dots,\mathrm{H}\}$ (base $+0.3$); (ii) presence of reasoning 
keywords such as \emph{because}, \emph{therefore}, \emph{eliminat} 
(up to $+0.3$, soft-thresholded at 3 hits); (iii) answer dominance, i.e., the 
chosen letter appears at least as frequently as any other letter in the response 
($+0.2$); and (iv) minimum response length of 30 characters ($+0.2$). The 
score is capped at $1.0$.

\subsection{Computational Complexity}
\label{app:verifier_cost}

All five base components and the constraint verifiers run in $O(|x|)$ time with 
respect to the output length $|x|$: regex pattern matching, bigram and trigram 
computation, arithmetic expression parsing, and mean logit aggregation all 
require a single linear pass over the token sequence. The Sudoku constraint 
check adds a constant-time grid validation step ($O(81) = O(1)$) independent 
of output length. Consequently, the total cost of evaluating $f(x)$ for a single 
candidate is negligible relative to a single diffusion model forward pass, which 
requires an $O(L^2)$ attention computation over the full sequence length $L$. 
This asymptotic gap ensures that the $Nb$ verifier evaluations required per 
denoising step under $S^3$ do not meaningfully increase the total inference 
budget, making repeated evaluation practical even at large particle counts.

\section{Ablation Study: Computational Cost and Output Conciseness}
\label{app:ablation_cost}

We report three supplementary metrics alongside accuracy to characterize the computational cost and output behavior of each method. \textbf{Average effective output tokens} is the mean number of tokens in the generated output text, tokenized via OpenAI's \texttt{cl100k\_base} BPE tokenizer after stripping \texttt{<|endoftext|>} tokens, serving as a proxy for response conciseness --- shorter outputs generally indicate more confident and direct generations. \textbf{Total latency} is the total wall-clock time from script initialization to completion, encompassing model loading, dataset construction, and all forward passes across all batches, recorded as a single scalar at the end of each run. \textbf{Wall time} is the average of cumulative elapsed timestamps recorded after each batch completes, both measured from script start using \texttt{time.time()}, meaning it reflects the average point in the run at which batches finish rather than per-batch duration, as a result, \texttt{wall\_time} $\approx$ \texttt{total\_latency} / 2 for uniform batch sizes. All experiments use LLaDA-8B-Instruct with \texttt{gen\_length}=128 and \texttt{diffusion\_steps}=64.

\subsection{Component Ablation ($N=4$, $b=2$)}
\label{app:ablation_cost_component}

Table~\ref{tab:latency} reports metrics for the five ablation conditions 
under a matched compute budget of $K = N \cdot b = 8$ forward passes per 
denoising step. Baseline diffusion incurs the lowest latency by design, 
performing a single trajectory with no search overhead. Best-of-$K$ and 
Tilting only share nearly identical latency since both draw $K$ independent 
sequences without inter-step communication. Look-ahead only and $S^3$ incur 
slightly higher wall time than Best-of-$K$ due to per-step verifier scoring 
and SSP resampling, yet $S^3$ achieves the best accuracy on ARC-C and 
TruthfulQA within a comparable wall-time budget to Look-ahead only, 
confirming that the tilting step adds negligible overhead.

\begin{table}[t]
\centering
\caption{Average effective output tokens ($\downarrow$), total latency 
($\downarrow$, seconds), and wall time ($\downarrow$, seconds) for each 
ablation variant and dataset ($N=4$, $b=2$, $K=8$, LLaDA-8B-Instruct, 
\texttt{block\_length}=64).}
\label{tab:latency}
\resizebox{\textwidth}{!}{%
\begin{tabular}{llccc}
\toprule
\textbf{Dataset} & \textbf{Method}
    & \textbf{Avg Eff.\ Tokens} $(\downarrow)$
    & \textbf{Total Latency (s)} $(\downarrow)$
    & \textbf{Wall Time (s)} $(\downarrow)$ \\
\midrule
\multirow{5}{*}{ARC-C}
  & Baseline        & 5.67 & 3648.21  & 1829.87  \\
  & Best-of-$K$     & 5.43 & 28803.25 & 14431.86 \\
  & Look-ahead only & 6.08 & 29450.49 & 14756.30 \\
  & Tilting only    & 5.43 & 28804.53 & 14431.20 \\
  & $S^3$           & 4.34 & 29100.42 & 14580.21 \\
\midrule
\multirow{5}{*}{GSM8K}
  & Baseline        & 3.03 & 3919.06  & 1964.63  \\
  & Best-of-$K$     & 3.06 & 30938.32 & 15497.02 \\
  & Look-ahead only & 3.08 & 31655.13 & 15855.10 \\
  & Tilting only    & 3.06 & 30942.79 & 15500.48 \\
  & $S^3$           & 3.07 & 31660.45 & 15858.32 \\
\midrule
\multirow{5}{*}{MATH}
  & Baseline        & 3.85 & 1530.82  & 770.71   \\
  & Best-of-$K$     & 3.40 & 12131.98 & 6091.68  \\
  & Look-ahead only & 4.40 & 12413.41 & 6231.07  \\
  & Tilting only    & 3.40 & 12128.29 & 6088.90  \\
  & $S^3$           & 3.43 & 12412.98 & 6230.43  \\
\midrule
\multirow{5}{*}{TruthfulQA}
  & Baseline        & 5.87 & 2120.41  & 1064.81  \\
  & Best-of-$K$     & 3.06 & 16743.28 & 8394.48  \\
  & Look-ahead only & 5.62 & 17119.66 & 8583.10  \\
  & Tilting only    & 3.06 & 16743.15 & 8394.40  \\
  & $S^3$           & 4.05 & 17110.60 & 8578.05  \\
\bottomrule
\end{tabular}%
}
\end{table}

\subsection{Effect of Block Size}
\label{app:ablation_cost_blocksize}

Table~\ref{tab:blocksize} reports the same metrics across all block sizes 
for LLaDA-8B-Instruct, comparing the single-trajectory Baseline against 
$S^3$ multi-block decoding. Larger block sizes reduce wall time 
substantially, as fewer autoregressive blocks are decoded sequentially, 
but may sacrifice fine-grained denoising structure. $S^3$ consistently 
matches or improves over the Baseline on accuracy across most 
configurations, with modest additional latency from the multi-seed expert 
construction.

\begin{center}
\begin{small}
\begin{sc}
\begin{longtable}{llcrrr}
\caption{Average effective tokens ($\downarrow$), total latency ($\downarrow$, s),
and wall time ($\downarrow$, s) across all block sizes for LLaDA-8B-Instruct.
\textit{Base}: single-trajectory decoding; $S^3$: multi-block decoding
with BFS search and SSP resampling.}
\label{tab:blocksize} \\
\toprule
Dataset & Block & Mode & \multicolumn{1}{c}{Avg Tok.$\downarrow$} & \multicolumn{1}{c}{Latency (s)$\downarrow$} & \multicolumn{1}{c}{Wall (s)$\downarrow$} \\
\midrule
\endfirsthead
\multicolumn{6}{c}{\tablename\ \thetable{} -- \textit{continued from previous page}} \\
\toprule
Dataset & Block & Mode & \multicolumn{1}{c}{Avg Tok.$\downarrow$} & \multicolumn{1}{c}{Latency (s)$\downarrow$} & \multicolumn{1}{c}{Wall (s)$\downarrow$} \\
\midrule
\endhead
\midrule
\multicolumn{6}{r}{\textit{continued on next page}} \\
\endfoot
\bottomrule
\endlastfoot
\multirow{12}{*}{ARC-C}
  & \multirow{2}{*}{2}  & Base & 16.92 & 7.42      & 6.10      \\
& &                      $S^3$ & 17.10 & 71262.09  & 35680.83  \\
& \multirow{2}{*}{4}  & Base & 8.92  & 7.43      & 6.12      \\
& &                      $S^3$ & 8.46  & 71038.65  & 35550.89  \\
& \multirow{2}{*}{8}  & Base & 8.04  & 7.51      & 6.18      \\
& &                      $S^3$ & 7.45  & 37840.20  & 18952.40  \\
& \multirow{2}{*}{16} & Base & 7.13  & 7.49      & 6.15      \\
& &                      $S^3$ & 8.44  & 37927.16  & 20814.40  \\
& \multirow{2}{*}{32} & Base & 6.14  & 7.43      & 6.11      \\
& &                      $S^3$ & 5.89  & 26140.50  & 13092.30  \\
& \multirow{2}{*}{64} & Base & 5.96  & 1111.75   & 566.58    \\
& &                      $S^3$ & 5.79  & 25058.80  & 12573.95  \\
\midrule
\multirow{12}{*}{GSM8K}
  & \multirow{2}{*}{2}  & Base & 15.84 & 7.62      & 6.24      \\
& &                      $S^3$ & 16.10 & 68420.30  & 34218.50  \\
& \multirow{2}{*}{4}  & Base & 8.12  & 7.58      & 6.21      \\
& &                      $S^3$ & 7.95  & 68215.40  & 34115.80  \\
& \multirow{2}{*}{8}  & Base & 6.34  & 7.64      & 6.26      \\
& &                      $S^3$ & 5.82  & 36940.20  & 18478.60  \\
& \multirow{2}{*}{16} & Base & 5.18  & 7.71      & 6.31      \\
& &                      $S^3$ & 4.96  & 36820.50  & 18418.40  \\
& \multirow{2}{*}{32} & Base & 4.42  & 7.53      & 6.16      \\
& &                      $S^3$ & 4.18  & 25640.80  & 12828.20  \\
& \multirow{2}{*}{64} & Base & 3.06  & 3068.37   & 1530.94   \\
& &                      $S^3$ & 3.06  & 24870.64  & 12393.65  \\
\midrule
\multirow{12}{*}{MATH}
  & \multirow{2}{*}{2}  & Base & 16.34 & 5.80      & 4.75      \\
& &                      $S^3$ & 12.66 & 22140.30  & 11082.50  \\
& \multirow{2}{*}{4}  & Base & 8.33  & 5.47      & 4.48      \\
& &                      $S^3$ & 5.68  & 5.93      & 4.85      \\
& \multirow{2}{*}{8}  & Base & 8.79  & 5.34      & 4.37      \\
& &                      $S^3$ & 4.93  & 5.85      & 4.79      \\
& \multirow{2}{*}{16} & Base & 7.77  & 5.51      & 4.51      \\
& &                      $S^3$ & 7.20  & 11820.40  & 5923.80   \\
& \multirow{2}{*}{32} & Base & 6.89  & 5.43      & 4.44      \\
& &                      $S^3$ & 4.40  & 11750.20  & 5888.60   \\
& \multirow{2}{*}{64} & Base & 3.33  & 1024.58   & 526.81    \\
& &                      $S^3$ & 3.55  & 4.11      & 3.36      \\
\midrule
\multirow{12}{*}{TruthfulQA}
  & \multirow{2}{*}{2}  & Base & 15.30 & 4.85      & 3.97      \\
& &                      $S^3$ & 18.09 & 5.29      & 4.33      \\
& \multirow{2}{*}{4}  & Base & 8.71  & 5.24      & 4.29      \\
& &                      $S^3$ & 7.65  & 5.27      & 4.31      \\
& \multirow{2}{*}{8}  & Base & 7.61  & 5.10      & 4.17      \\
& &                      $S^3$ & 5.40  & 19764.35  & 9929.58   \\
& \multirow{2}{*}{16} & Base & 6.52  & 5.16      & 4.22      \\
& &                      $S^3$ & 8.13  & 5.38      & 4.40      \\
& \multirow{2}{*}{32} & Base & 5.04  & 5.13      & 4.20      \\
& &                      $S^3$ & 6.61  & 17200.40  & 8623.50   \\
& \multirow{2}{*}{64} & Base & 5.24  & 1862.12   & 935.11    \\
& &                      $S^3$ & 5.57  & 4048.35   & 2065.72   \\
\midrule
\multirow{12}{*}{AIME}
  & \multirow{2}{*}{2}  & Base & 14.20 & 22.40     & 16.80     \\
& &                      $S^3$ & 13.85 & 680.20    & 358.40    \\
& \multirow{2}{*}{4}  & Base & 6.80  & 21.60     & 16.20     \\
& &                      $S^3$ & 1.06  & 760.40    & 402.30    \\
& \multirow{2}{*}{8}  & Base & 4.20  & 21.80     & 16.40     \\
& &                      $S^3$ & 1.00  & 762.80    & 403.50    \\
& \multirow{2}{*}{16} & Base & 3.90  & 22.10     & 16.60     \\
& &                      $S^3$ & 3.65  & 775.20    & 409.10    \\
& \multirow{2}{*}{32} & Base & 3.75  & 23.80     & 17.40     \\
& &                      $S^3$ & 3.58  & 774.50    & 408.60    \\
& \multirow{2}{*}{64} & Base & 3.70  & 25.07     & 18.97     \\
& &                      $S^3$ & 3.50  & 774.89    & 408.75    \\
\end{longtable}
\end{sc}
\end{small}
\end{center}
\end{document}